\documentclass[journal]{IEEEtran}
\usepackage{cite}
\usepackage[switch,pagewise]{lineno}
\usepackage{url,subfigure}
\usepackage[hidelinks,hypertexnames=false]{hyperref}
\usepackage{booktabs}
\usepackage{multirow}
\usepackage{amssymb}
\usepackage{graphicx}
\usepackage{algorithm}
\usepackage{algpseudocode}
\usepackage{amsmath}
\usepackage{makecell}
\usepackage{ulem}
\usepackage{color}
\usepackage[table]{xcolor}
\usepackage{cuted}
\usepackage{capt-of}
\usepackage{pifont}
\usepackage{marvosym}

\usepackage{scalerel}
\usepackage{tikz}
\usetikzlibrary{svg.path}
\definecolor{orcidlogocol}{HTML}{A6CE39}
\tikzset{
  orcidlogo/.pic={
    \fill[orcidlogocol] svg{M256,128c0,70.7-57.3,128-128,128C57.3,256,0,198.7,0,128C0,57.3,57.3,0,128,0C198.7,0,256,57.3,256,128z};
    \fill[white] svg{M86.3,186.2H70.9V79.1h15.4v48.4V186.2z}
                 svg{M108.9,79.1h41.6c39.6,0,57,28.3,57,53.6c0,27.5-21.5,53.6-56.8,53.6h-41.8V79.1z M124.3,172.4h24.5c34.9,0,42.9-26.5,42.9-39.7c0-21.5-13.7-39.7-43.7-39.7h-23.7V172.4z}
                 svg{M88.7,56.8c0,5.5-4.5,10.1-10.1,10.1c-5.6,0-10.1-4.6-10.1-10.1c0-5.6,4.5-10.1,10.1-10.1C84.2,46.7,88.7,51.3,88.7,56.8z};
  }
}
\newcommand\orcidicon[1]{\href{https://orcid.org/#1}{\mbox{\scalerel*{
\begin{tikzpicture}[yscale=-1,transform shape]
\pic{orcidlogo};
\end{tikzpicture}
}{|}}}}
\usepackage{hyperref}
\hypersetup{
    colorlinks=true,
    linkcolor=blue,
    filecolor=black,      
    urlcolor=black,
    citecolor=purple
}
\usepackage{tikz-network}
\newcommand{\egi}{\textit{e.g.}}
\newcommand{\settablefont}{\fontsize{6.9}{11.8}\selectfont}

\newcommand\clb[1]{{\color{blue}{#1}}}

\definecolor{SYellow}{RGB}{150,50,10}

\begin{document}
\normalem
\title{
\huge DST-Calib: A Dual-Path, Self-Supervised, Target-Free LiDAR-Camera Extrinsic Calibration Network
}

\author{
Zhiwei Huang,
Yanwei Fu,
Yi Zhou,
Xieyuanli Chen,
Qijun Chen,
and Rui Fan\textsuperscript{\Letter}

\thanks{Zhiwei Huang is with the Department of Control Science \& Engineering, the College of Electronics \& Information Engineering, Tongji University, Shanghai 201804, China (e-mail: zhiwei.huang@outlook.com).}
\thanks{Yanwei Fu is with the Department of Mechanical Engineering, the School of Mechanical Engineering, Tongji University, Shanghai 201804, China (e-mail: 2350435@tongji.edu.cn)}
\thanks{Yi Zhou is with the School of Robotics, Hunan University 410082, Changsha, China (e-mail: eeyzhou@hnu.edu.cn).}
\thanks{Xieyuanli Chen is with the College of Intelligence Science and Technology, National University of Defense Technology, Changsha 410073, China (e-mail: chenxieyuanli@hotmail.com).}
\thanks{Rui Fan is with the Department of Control Science \& Engineering, the College of Electronics \& Information Engineering, Shanghai Research Institute for Intelligent Autonomous Systems, the State Key Laboratory of Intelligent Autonomous Systems, and Frontiers Science Center for Intelligent Autonomous Systems, Tongji University, Shanghai 201804, China, as well as with the National Key Laboratory of Human-Machine Hybrid Augmented Intelligence, Institute of Artificial Intelligence and Robotics, Xi'an Jiaotong University, Xi'an 710049, Shaanxi, China (e-mail: rui.fan@ieee.org).}
\thanks{{\Letter} Corresponding author: Rui Fan.}
}

\maketitle	

\begin{abstract}
LiDAR-camera extrinsic calibration is essential for multi-modal data fusion in robotic perception systems. However, existing approaches typically rely on handcrafted calibration targets (\textit{e.g.}, checkerboards) or specific, static scene types, limiting their adaptability and deployment in real-world autonomous and robotic applications.
This article presents the first self-supervised LiDAR-camera extrinsic calibration network that operates in an online fashion and eliminates the need for specific calibration targets. We first identify a significant generalization degradation problem in prior methods, caused by the conventional single-sided data augmentation strategy. To overcome this limitation, we propose a novel double-sided data augmentation technique that generates multi-perspective camera views using estimated depth maps, thereby enhancing robustness and diversity during training.
Built upon this augmentation strategy, we design a dual-path, self-supervised calibration framework that reduces the dependence on high-precision ground truth labels and supports fully adaptive online calibration. Furthermore, to improve cross-modal feature association, we replace the traditional dual-branch feature extraction design with a difference map construction process that explicitly correlates LiDAR and camera features. This not only enhances calibration accuracy but also reduces model complexity.
Extensive experiments conducted on five public benchmark datasets, as well as our own recorded dataset, demonstrate that the proposed method significantly outperforms existing approaches in terms of generalizability. 
\end{abstract}

\section{Introduction}
\label{sec.intro}

Robots with human-like intelligence have long been envisioned as a central aspiration of robotics research \cite{giulia2025intergrat}. Today, this vision is becoming increasingly attainable with the rapid advancement of multi-modal sensor fusion systems. By integrating complementary information from diverse sensors, these systems significantly enhance robot perception, enabling reliable execution of complex tasks in real-world environments \cite{cattaneo2025cmrnext}. Among the most widely used sensors are LiDARs and cameras: the former provide precise geometric and spatial measurements, while the latter capture rich semantic and textural details \cite{zheng2025fastlivo2}. When fused, the complementary strengths of these modalities enable robots to achieve robust environmental understanding \cite{li2022deepfusion}. 
This enhanced perception capability plays a crucial role in supporting reliable performance across a range of fundamental robotic tasks, including odometry \cite{tian2022kimera}, object recognition \cite{zhang2024toward}, and localization \cite{luo2025bevplace++}.

\begin{figure}
    \centering
    \includegraphics[width=0.985\linewidth]{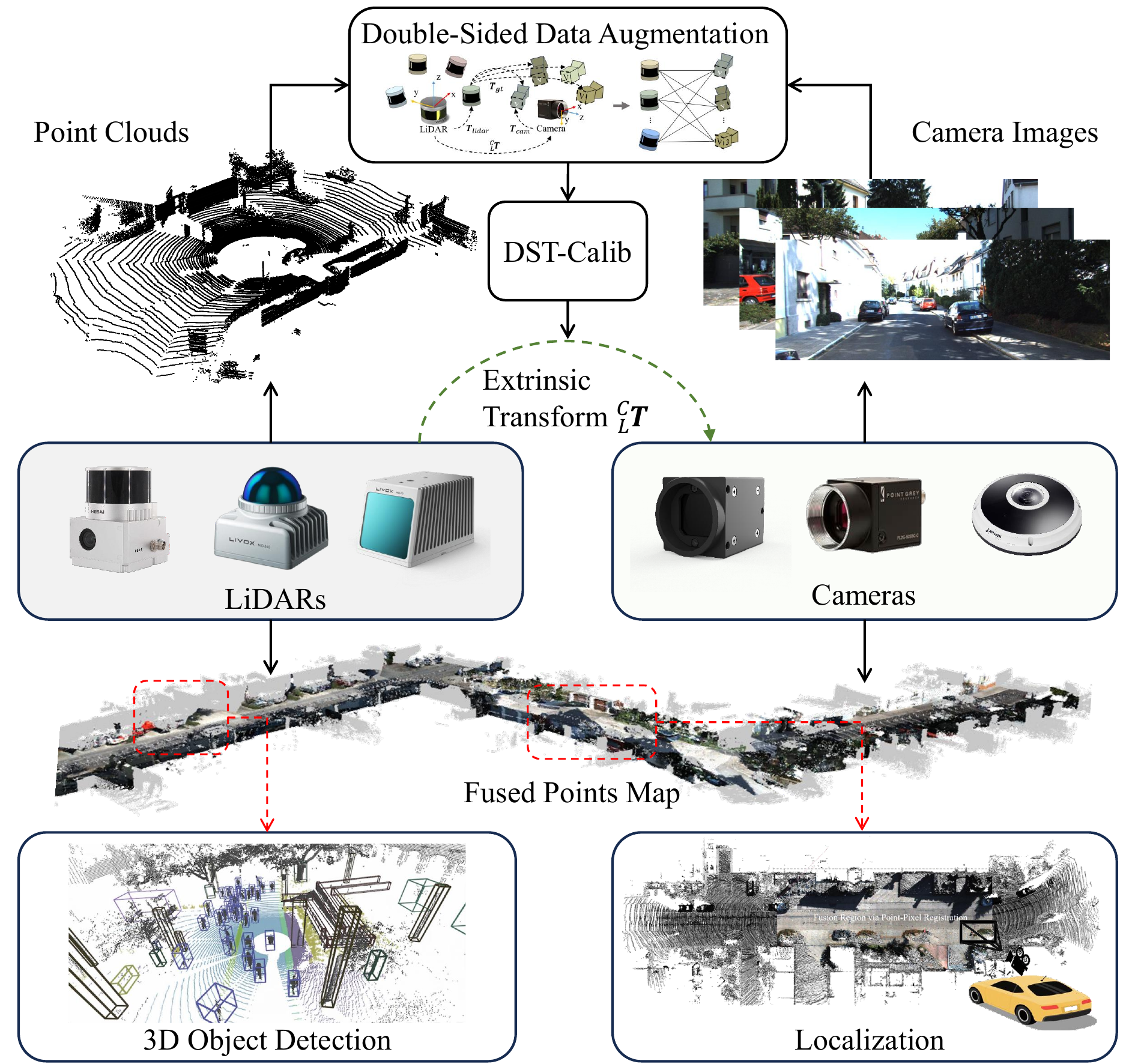}
    \caption{Our proposed DST-Calib estimates the extrinsic transformation with six degrees of freedom (6-DoF) between LiDAR scans and camera images in the wild. It can be readily employed for robotic tasks such as object detection, odometry, and localization.}
    \label{fig.cover}
\end{figure}

As depicted in Fig. \ref{fig.cover}, LiDAR-camera extrinsic calibration (LCEC), which estimates the extrinsic transformation between the two sensors, is the core and foundational process for effective data fusion. 
Existing LCEC methods are generally categorized as either target-based or target-free, depending on whether the algorithm relies on pre-defined calibration targets. Target-based calibration approaches provide stable and reliable performance in static and controlled environments where calibration targets (\textit{e.g.}, checkerboards or customized 3D pattern boards) are available \cite{cui2020acsc,koo2020analytic}. However, with the increasing demand for dynamic applications (such as autonomous mobile robots and multi-robot collaborative perception), where extrinsic parameters may fluctuate due to vibrations, mechanical shocks, or active sensor movement, target-based approaches become impractical and often fail to maintain valid calibration for reliable data fusion. 
Although a variety of traditional target-free LCEC methods have been proposed to eliminate the reliance on specific calibration targets by leveraging cross-modal geometric textures or mutual information to align LiDAR point clouds and camera images  \cite{yuan2021pixel,zhu2020online,pandey2015automatic}, they still struggle to adapt to complex and unstructured real-world scenes and often lack sufficient accuracy, leaving considerable room for improvement.

\begin{figure}[t!]
    \centering
    \includegraphics[width=0.985\linewidth]{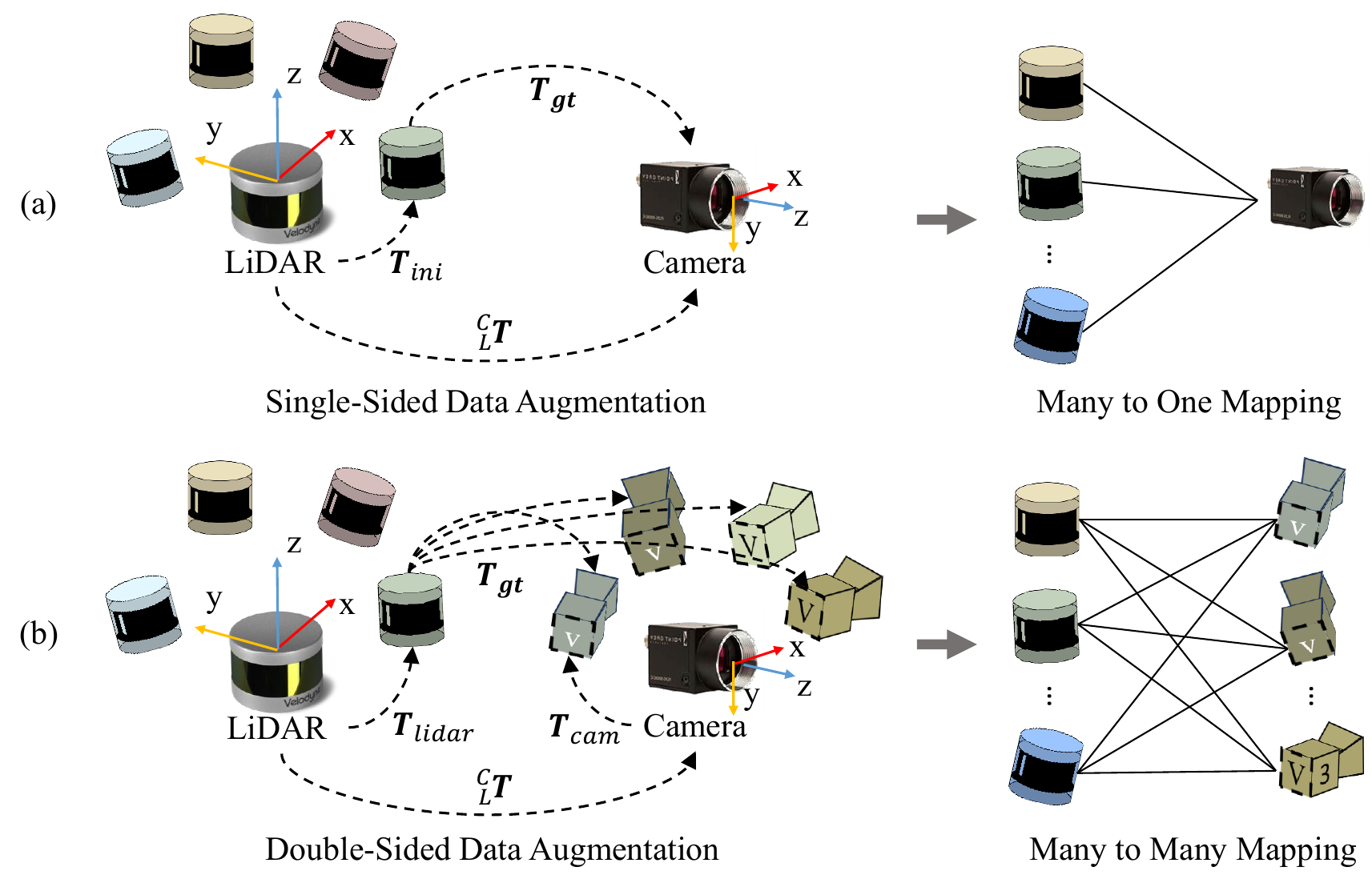}
    \caption{Mapping relationship of different data augmentation methods in the training process of LCEC network: (a) the classical single-sided data augmentation with many-to-one mapping; (b) our proposed double-sided approach with many-to-many mapping.}
    \label{fig.data_augmentation_eg}
\end{figure}

Advances in deep learning techniques have driven significant exploration into enhancing the accuracy and efficiency of traditional target-free algorithms. 
Researchers attempt to design end-to-end deep neural networks (DNN) \cite{borer2024chaos, lv2021lccnet,shi2020calibrcnn,zhao2021calibdnn,iyer2018calibnet,yuan2020rggnet} that directly regress 6-DoF extrinsic parameters with the sensor inputs. These DNN-based methods have the advantage of being easy to train and usually achieve high precision when evaluated on images captured with the same camera used during training. However, they have the major drawback of being dependent on the specific camera’s parameters. 
In this case, end-to-end DNN models trained on a single LiDAR-camera pair typically tend to overfit to the specific intrinsic and extrinsic parameters of the training dataset and fail to generalize even when evaluated on a different camera in the same dataset.  
Additionally, we observe an inherent data imbalance that biases the network toward relying almost exclusively on LiDAR point clouds for extrinsic prediction. In conventional training data preparation, the camera position is typically fixed, and prior works adopt a single-sided data augmentation strategy that introduces perturbations solely around the camera’s pose. As illustrated in Fig. \ref{fig.data_augmentation_eg}, each projected point cloud corresponds to exactly one camera image, thus establishing a many-to-one mapping. Under this mapping relationship, the camera image branch of the DNN gradually deteriorates during training. The fixed relative pose relationship causes the network to overfit to simplistic, redundant features commonly present in LiDAR projections, leading it to memorize the extrinsic parameters rather than learning meaningful, robust cross-modal correlations. As a result, the trained model becomes highly dependent on the specific camera used during training and fails to generalize across different sensor configurations, requiring retraining whenever the camera is changed. 

In this article, we aim to address these significant issues that exist in the previous DNN-based LCEC approach and improve the generalization ability of current target-free approaches. First, we show that if camera positions are diversified through a double-sided data augmentation (which ensures that a single point cloud projection maps to multiple camera images, forming a many-to-many relationship), the same network architecture achieves significantly improved generalization. 
To build this double-sided data augmentation, we employ monocular depth estimation to reconstruct camera depth point clouds from raw camera images. An efficient depth correction algorithm, \textbf{D}epth \textbf{A}nchor \textbf{R}efinement (DAR), is designed to correct the initial depth estimation with the sparse guidance of the LiDAR point clouds. By converting the corrected camera depth map into 3D depth clouds, we can generate high-quality depth projections from different perspective views on both the LiDAR and camera sides, thereby achieving balanced double-sided data augmentation. 
Moreover, to address the heavy dependence on ground-truth extrinsic parameters, we propose DST-Calib, the first target-free LCEC network that supports self-supervised learning. Unlike previous networks, DST-Calib contains dual pathways: a fully-supervised pathway and a self-supervised pathway. 
When ground truth is unavailable, DST-Calib can perform coarse 6-DoF extrinsic calibration or high-precision rotation-only calibration through online self-supervised learning. 
When ground truth labels are available, DST-Calib can achieve accurate and robust calibration through fully-supervised training and can easily generalize to unseen environments.
In addition, we demonstrate that the classical double-branch architecture of prior arts, which comprises two independent feature extraction branches for LiDAR and camera inputs, is not essential for achieving high calibration accuracy.
By constructing difference maps between LiDAR and camera depth projections, we explicitly associate LiDAR and camera data, thereby cutting off a feature extraction branch and building a single-branch architecture.  
Despite containing fewer model parameters, this single-branch method achieves better performance than the conventional double-branch architecture. In addition, extensive experiments across five public datasets, covering a total of 23 distinct sensor configurations, demonstrate the superior performance of DST-Calib. A model trained on only 3 sensor configurations can directly generalize to all other sensor setups without finetuning, which greatly improves the generalization ability compared to the previous target-free LCEC.
Furthermore, most existing public datasets employ mechanical spinning LiDARs, while few provide dense 4D point clouds captured by solid-state LiDARs with repeatable scans. To facilitate research in this direction, we construct a new public dataset containing extensive pairs of dense point clouds and camera images, comprising 5 sequences recorded across diverse indoor and outdoor environments.

To summarize, our novel contributions are as follows:

\begin{itemize}
    \item{We revisit previous DNN-based LCEC networks and discover the large generalization degradation caused by the unbalanced single-sided training data augmentation. A novel double-sided training data augmentation strategy is developed to address this generalization problem.
    }
    \item{DAR is designed to refine the depth of monocular camera images with the guidance of the LiDAR point clouds. 
    }
    \item {DST-Calib, the first LCEC network that supports self-supervised learning, which is independent of any sensor-specific parameter. 
    }
    \item {Extensive experiments on five public datasets demonstrate the generalization problem of the single-sided data augmentation, the necessity of our proposed double-sided data augmentation, and the state-of-the-art (SoTA) performance of DST-Calib. In addition, we record a new real-world dataset, LCScenes, designed to support the training and evaluation of LCEC networks on dense point clouds captured by solid-state LiDARs.}
\end{itemize}

\section{Related Work}
\label{sec.related_work}
\begin{figure*}[t!]
    \centering
    \includegraphics[width=0.99\linewidth]{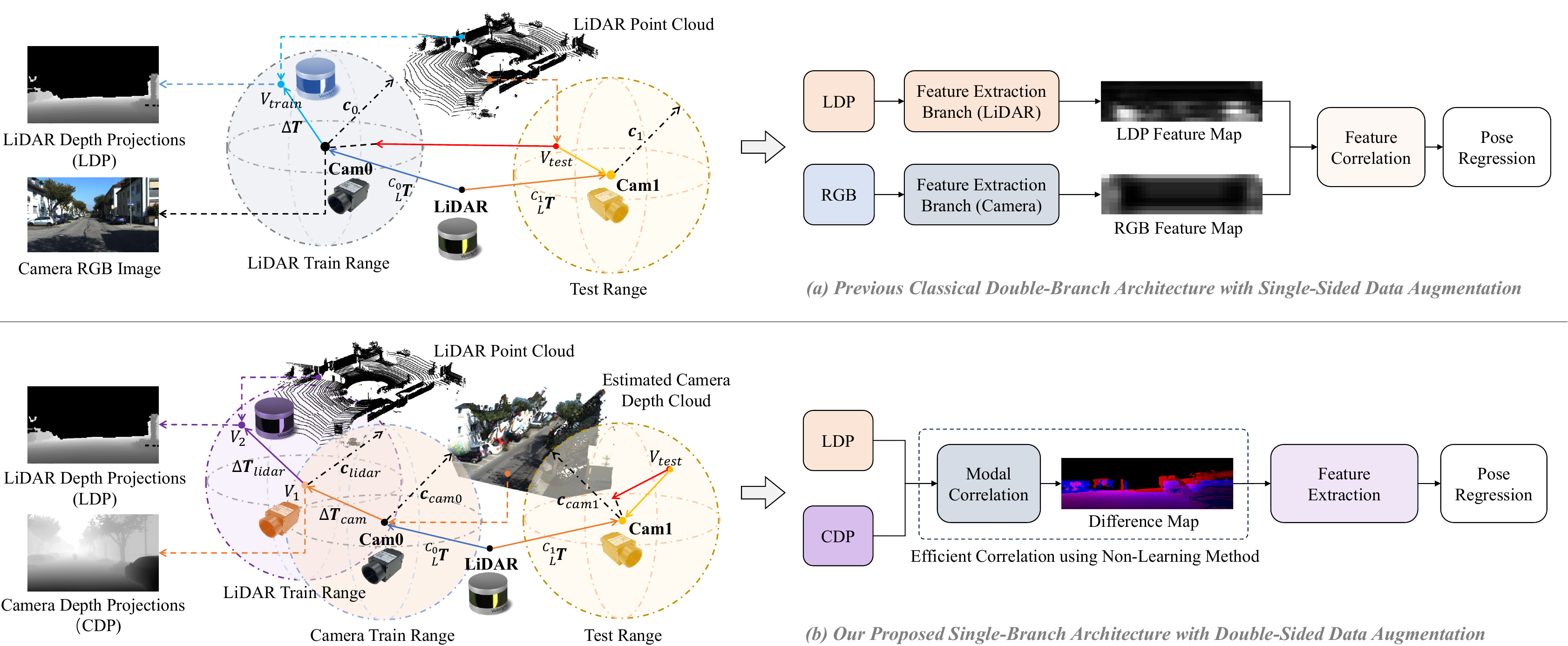}
    \caption{The comparison of the DNN-based LiDAR-camera extrinsic calibration pipelines: (a) Classical double-branch architecture with single-sided data augmentation; (b) Our proposed single-branch architecture with double-sided data augmentation.}
    \label{fig.two_network_architecture}
\end{figure*}

\subsection{Traditional Target-Free Approaches}
\label{sec.related_work_traditional_target_free}
To eliminate the need for specific calibration targets and the expense of offline data acquisition and optimization, researchers have begun developing target-free approaches that can perform extrinsic calibration in sensor operating environments. In early attempts, studies \cite{lv2015automatic, yuan2021pixel, pandey2015automatic} estimate the relative pose by matching cross-modal edges or mutual information extracted from LiDAR point clouds and camera images. Although these methods are effective in some scenarios where geometric and textural features are abundant, they rely heavily on uniformly distributed lines and rich texture details, which significantly reduce robustness. 
Recent progress in deep learning has motivated extensive efforts to improve target-free calibration methods. A number of works \cite{li2018automatic, ma2021crlf, wang2022automatic, han2021auto, liao2023se, zhu2020online, koide2023general, zhiwei2024lcec} incorporate learning-based tools into calibration pipelines to boost both robustness and efficiency. For example, \cite{ma2021crlf} achieves calibration by matching lane markings and pole-like structures extracted through semantic segmentation. In a similar vein, \cite{han2021auto} treats stop signs as semantic landmarks and progressively refines the calibration parameters using a Kalman filtering scheme. More recently, the study \cite{koide2023general} proposed Direct Visual LiDAR Calibration (DVL), a point-level approach that leverages SuperGlue \cite{sarlin2020superglue} to directly construct 3D-2D correspondences across LiDAR and camera modalities.

\subsection{DNN-Based Target-Free Approaches}
\label{sec.related_work_dnn_target_free}
In recent years, end-to-end DNN-based approaches have been proposed to estimate the extrinsic parameters between LiDAR and camera
sensors. Previous DNN-based LCEC networks primarily follow the pioneering work, RegNet \cite{schneider2017regnet}. As illustrated in Fig.~\ref{fig.two_network_architecture}(a), these methods adopt a similar double-branch architecture consisting of dual feature extraction branches, a feature correlation module, and a pose regression module. Subsequent works, such as CalibNet \cite{iyer2018calibnet}, RGGNet \cite{yuan2020rggnet}, CalibDNN \cite{zhao2021calibdnn}, LCCNet \cite{lv2021lccnet}, and CalibDepth \cite{zhu2023calibdepth}, follow this double-branch architecture and propose different network modules to improve the performance. 

During training, these DNN-based approaches adopt a similar single-sided data augmentation method. LiDAR projections from different perspectives around the camera side are generated to simulate the calibration initial estimate \cite{lv2021lccnet}. The underlying assumption of the single-sided data augmentation strategy is that, by projecting depth images within a sufficiently large mis-calibration range from the camera center, the network can learn the correspondence between LiDAR point clouds and camera images, thereby enabling accurate estimation of the extrinsic parameters. However, this single-sided data augmentation establishes a many-to-one mapping relationship of the input LiDAR projections and camera images. Although the LiDAR projections are captured from random perspectives relative to the camera, each projection corresponds to only one relative pose. As a result, the model can easily infer the extrinsic transformation solely from the features of the LiDAR projection, without the need to incorporate any information from the camera data. However, for a calibration task, the camera position is uncertain. The LiDAR projections and camera images should establish a many-to-many mapping. In other words, each LiDAR projection might correspond to an infinite number of possible camera images. We believe that only a model trained under this assumption can possibly have generalization ability to other sensor configurations.  

Additionally, we visualize the feature maps produced by the double-branch network trained with the conventional single-sided data augmentation strategy. As illustrated in Fig. \ref{fig.two_network_architecture}, the feature map derived from the camera branch appears almost entirely black, containing little to no informative content. For further validation, we replace the RGB input with an all-zero image in the training. As expected, the trained model was still able to predict extrinsic parameters accurately within the training dataset. This further proves that the current training strategy is fundamentally misleading, as the camera branch does not contribute to the extrinsic estimation.  
Paradoxically, this phenomenon also suggests that, for regressing an extrinsic matrix, a single branch is sufficient to achieve competitive results. In fact, the double-branch architecture may even hinder learning effective content, as it increases the difficulty of feature correlation due to the substantial modality gap between LiDAR point clouds and camera images.  

In summary, existing LCEC networks mainly suffer from two critical issues:  
\begin{enumerate}
    \item Models trained with the misleading single-sided data augmentation method can only estimate extrinsic parameters specific to a single LiDAR-camera setup and lack generalization to other configurations.  
    \item The double-branch architecture is not only unnecessary for feature extraction and correlation, but also increases the challenge of cross-modal feature correlation.  
\end{enumerate}

To address these two critical issues, we propose a novel double-sided data augmentation strategy and a dual-path network architecture that supports self-supervised learning. 
Unlike previous DNN-based methods, our DST-Calib is no longer limited to specific sensor parameters, which can be directly applied to diverse challenging real-world scenarios to achieve online, target-free calibration.

\section{Methodology}
\label{sec.method}

In this section, we begin by presenting our novel training strategy based on double-sided data augmentation. Next, we introduce our proposed novel calibration framework, which supports both fully-supervised and self-supervised learning. Finally, we detail the novel designs of DST-Calib, which includes a single-branch network architecture that directly estimates the 6-DoF extrinsic transformation by analyzing the difference map constructed from the LiDAR point cloud and the camera depth cloud.

\subsection{Double-Sided Data Augmentation}
\label{sec.method_current_problems}
The goal of LCEC is to estimate the extrinsic transformation
\begin{equation}
{^{C}_{L}\boldsymbol{T}} = 
\begin{pmatrix}
{^{C}_{L}\boldsymbol{R}} & {^{C}_{L}\boldsymbol{t}} \\
\boldsymbol{0}^\top & 1
\end{pmatrix}
\in{SE(3)},
\label{eq.lidar_to_camera_trans}
\end{equation}
between the LiDAR and the camera. $^{C}_{L}\boldsymbol{R} \in{SO(3)}$ represents the rotation matrix, $^{C}_{L}\boldsymbol{t}$ denotes the translation vector, and $\boldsymbol{0}$ represents a column vector of zeros. 

Previous DNN-based LCEC networks adopt a classical single-sided data augmentation strategy (as depicted in Fig. \ref{fig.data_augmentation_eg}(a) and Fig.~\ref{fig.two_network_architecture}(a)) to construct training samples. Specifically, they generate misaligned point cloud projections using $\Delta{\boldsymbol{T}}$ within a limited perturbation range around the camera. For every $\Delta{\boldsymbol{T}}$, there is a corresponding initial extrinsic transformation 
\begin{equation}
\boldsymbol{T}_{ini} = \Delta{\boldsymbol{T}}(^C_L\boldsymbol{T})
\label{eq.singel_sided_aug}
\end{equation}
to decide where to project the LiDAR depth projection (LDP) images. Each generated LDP image corresponds directly to the RGB (or grayscale) image captured from the same camera viewpoint, forming the LDP-RGB training pairs. This many-to-one data mapping relationship will mislead the network to solely depend on the input of the LiDAR-branch because LiDAR input is necessary enough to ``guess'' out the extrinsic transformation between the two sensors. The model trained under this data augmentation strategy will confront a huge generalization degradation when applied to the other sensor configurations, which are different from the training datasets.

To fundamentally overcome this generalization issue, we propose a novel double-sided augmentation framework (as depicted in Fig.~\ref{fig.two_network_architecture}(b)). Unlike the single-sided approach defined in \eqref{eq.singel_sided_aug} that restricts the augmentation to the LiDAR side, this double-sided method has to ensure that each LiDAR projection corresponds to multiple camera data, establishing a many-to-many mapping.
To realize this many-to-many mapping, we leverage monocular depth estimation (Depth Anything V2 \cite{depth_anything_v2} and MoGe2 \cite{wang2025moge}) to produce camera depth projection (CDP) images ${{^C_D}\boldsymbol{I}} \in \mathbb{R}^{H\times W}$ from raw camera images ${{^C_I}\boldsymbol{I}} \in \mathbb{R}^{3\times H\times W}$ and further design a depth refinement algorithm to obtain reliable depth clouds $\mathcal{Q} = \{[x_i,y_i,z_i]^\top\}^M_{i = 1}$ for generating projections from arbitrary poses. A series of initial extrinsic transforms is configured for both the LiDAR and camera sides:
\begin{equation}
\begin{aligned}
\boldsymbol{T}_{cam} &= \Delta{\boldsymbol{T}_{cam}}(^C_L\boldsymbol{T}), \\
\boldsymbol{T}_{lidar} &= \Delta{\boldsymbol{T}_{lidar}}(\boldsymbol{T}_{cam})
\end{aligned}
\label{eq.double_sided_aug}
\end{equation}
at two separate data augmentation centers, respectively.  The CDP images will be captured at $\boldsymbol{T}_{cam}$ within a mis-calibration range $\boldsymbol{c}_{cam}$ originating from the camera center. The LiDAR depth projection (LDP) images ${{^L_D}\boldsymbol{I}}$ will be generated using the pose $\boldsymbol{T}_{lidar}$ within a deviation range $\boldsymbol{c}_{lidar}$ originating at the positions of the generated CDP images. $\Delta{\boldsymbol{T}_{cam}}$ and $\Delta{\boldsymbol{T}_{lidar}}$ are randomly sampled inside the region decided by the mis-calibration range $\boldsymbol{c}_{cam}$ and $\boldsymbol{c}_{lidar}$. The goal of the calibration network is to estimate the relative extrinsic transform $\boldsymbol{T}_{gt}$ between a virtual LiDAR (denoted as $lidar$) positioned at $\boldsymbol{T}_{lidar}$ and a virtual camera (denoted as $cam$) at $\boldsymbol{T}_{cam}$, which is defined as follows:
\begin{equation}
{\boldsymbol{T}_{gt} = {\boldsymbol{T}_{cam}}({\boldsymbol{T}_{lidar}})^{-1}}.
\end{equation}
This approach establishes a true many-to-many mapping and substantially boosts data diversity by ensuring that: (1) each LiDAR point cloud projection corresponds to multiple camera perspectives, and (2) every camera image is paired with diverse LiDAR sampling configurations. Under this augmentation scheme, the network cannot simply infer extrinsic parameters from a single modality alone, thereby enforcing the learning of effective cross-modal contextual relationships.

\subsection{Depth Anchor-Based Efficient Depth Refinement}
\label{sec.method_dar}

Monocular depth estimation often fails to provide accurate metric depth in many scenarios. To obtain a high-quality depth map and ensure double-sided data augmentation, we propose DAR, a depth refinement method that leverages depth anchors derived from sparse LiDAR point clouds to correct initial depth maps estimated from raw camera images. This process reconstructs a dense depth map with reliable metric accuracy. The detailed procedure of DAR is described as follows:

\subsubsection{Piecewise Linear Depth Refinement via Anchor Points}
From each pair of LDP and CDP, we can extract a set of $K$ depth anchors 
\begin{equation}
\mathcal{A}=\{({d}^{C}_0,d^{L}_0),\,({d}^{C}_1,{d}^{L}_1),\dots,({d}^{C}_{K-1},{d}^{L}_{K-1})\},
\end{equation}
where ${d}^{C}_i\in[0,1]$ denotes the normalized monocular camera depth and
\({d}^{L}_i\in\mathbb{R}_+\) represents the corresponding LiDAR depth with accurate metric.
We perform a simple yet effective piecewise linear remapping $f\colon [0,1]\to\mathbb{R}_+$ of the entire estimated depth map ${^C_D}\hat{\boldsymbol{I}}_0$ by connecting adjacent anchors with straight line segments:
\begin{equation}
f({d}^{C})\! = \!\!
\begin{cases}
{d}^{L}_0, & {d}^{C}\le {d}^{C}_0,\\
{d}^{L}_{i-1} \!+\! \displaystyle\frac{{d}^{L}_i - {d}^{L}_{i-1}}{{d}^{C}_i - {d}^{C}_{i-1}} \bigl({d}^{C} - {d}^{C}_{i-1}\bigr), \!\!\!\!
& {d}^{C}_{i-1} < {d}^{C} \le {d}^{C}_i,\\
{d}^{L}_{K-1}, & {d}^{C} > {d}^{C}_{K-1}.
\end{cases}
\label{eq.reconstruct_depth_by_anchors}
\end{equation}
where $i=1,\dots,K-1$. In practice, the first and last segments will be projected onto constant values outside the anchor range if needed.

After establishing the piecewise linear depth refinement strategy in \eqref{eq.reconstruct_depth_by_anchors}, given the estimated depth ${^C_D}\hat{\boldsymbol{I}}_0(u,v)$ of the initial estimated depth map at pixel $\boldsymbol{p}$, the rectified depth can be obtained as follows:
\begin{equation}
\label{eq:depth-rectified}
{^C_D}\hat{\boldsymbol{I}}(u,v) = f\bigl({^C_D}\hat{\boldsymbol{I}}_0(u,v)\bigr).
\end{equation}
This approach preserves the monotonic ordering of depth while locally adapting it to anchor-derived LiDAR depths. The simple mapping ensures global depth ordering consistency while linearly calibrating within anchor bins, making it computationally efficient and easy to implement. It is especially well-suited for real-time applications or as a post-processing refinement step. While the depth estimation does not always retain the same monotonic, piecewise-linear behavior at certain isolated foreground objects (\egi, tree trunks, or pedestrians) as in other regions, we believe that DAR’s overall depth refinement quality is sufficient for multi-view data augmentation.

\subsubsection{Monotone, Near-Linear Anchor Selection}
\begin{figure}
    \centering
    \includegraphics[width=0.987\linewidth]{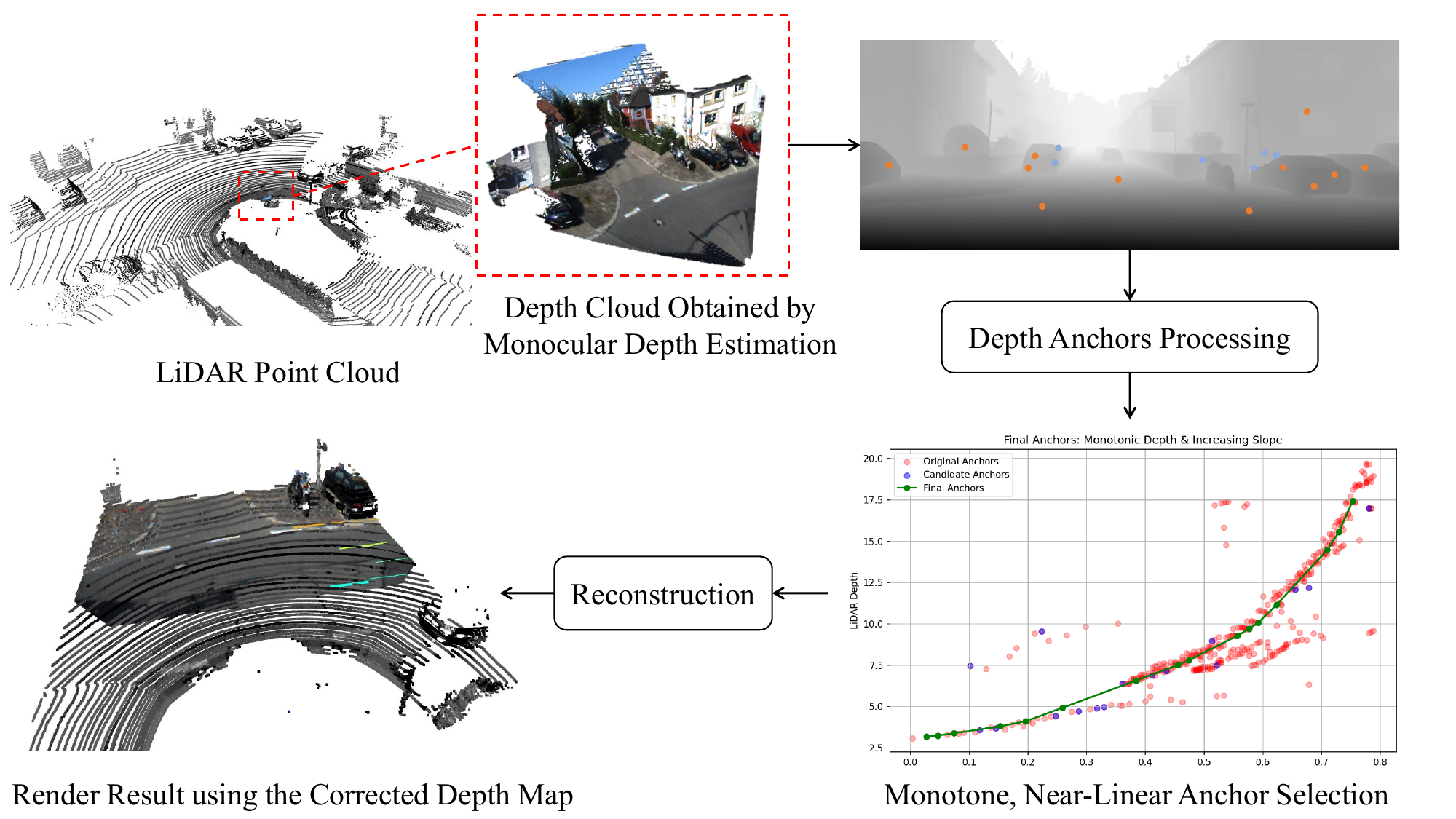}
    \caption{Depth refinement via LiDAR-camera depth anchor reconstruction.}
    \label{fig.depth_correction}
\end{figure}

\begin{figure}[t!]
    \centering
    \includegraphics[width=0.985\linewidth]{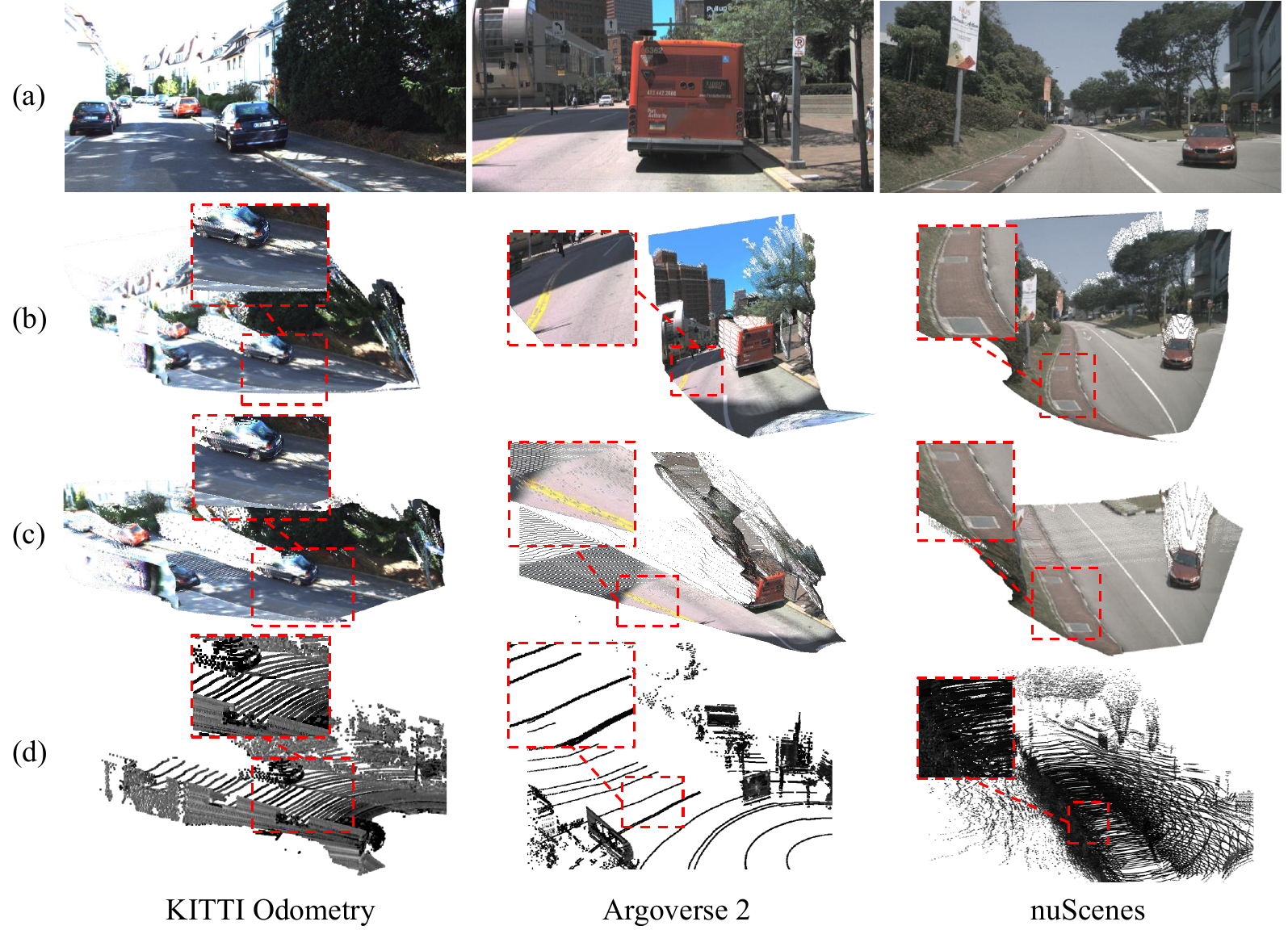}
    \caption{Qualitative comparisons between the initial estimated depth cloud and those after the correction of our DAR on various datasets: (a)-(d) RGB images, initial estimated depth cloud, corrected depth cloud, and corresponding point cloud scanned by LiDARs.}
    \label{fig.depth_correction_visualization}
\end{figure}

According to the depth reconstruction method (\ref{eq.reconstruct_depth_by_anchors}), if the depth anchors are accurate enough, DAR will probably obtain a much more reliable depth map than the initial one obtained by monocular depth estimation. So the key problem is to obtain reliable and accurate depth anchors. To tackle this, as depicted in Fig. \ref{fig.depth_correction}, DAR first sorts the estimated camera depth and LiDAR depth from small to large, and selects the depth value from the same coordinates of the two depth projections as initial depth anchors. Actually, these depth anchors are correspondences that are selected directly using the depth projection locations. Considering the uncertain reliability of the initial depth anchors, we use a monotone, near-linear anchor selection to filter out the unreliable anchors and utilize the more reliable ones to reconstruct a better depth map. 

In this depth anchor selection, we are given a set of noisy depth anchor correspondences $\mathcal{A}=\{(d^C_i,d^L_i)\}_{i=1}^{N}$, where $d^C_i \in [0,1]$ are estimated normalized depth from the camera image and $d^L_i \in \mathbb{R}_{+}$ are LiDAR depth. Let $\tilde{\mathcal{A}}=\{(d^C_{(i)},d^L_{(i)})\}_{i=1}^{N}$ denote $\mathcal{A}$ sorted by $d^C$, where $d^C_{(1)}\le \cdots \le d^C_{(N)}$. The goal is to select an ordered subsequence of anchors $\mathcal{S}=\{(d^C_k,d^L_k)\}_{k=1}^{K}$ with $K$ as large as possible, such that:
(1) it is monotone nondecreasing in both $d^C$ and $d^L$,
and (2) its local slopes are nondecreasing to avoid oscillations,
yielding a near-linear, shape-stable calibration curve.

For consecutive points in $\mathcal{S}$,  their secant slopes are defined as follows:
\begin{equation}
s_k \ :=\ \frac{d^L_{k+1}-d^L_k}{d^C_{k+1}-d^C_k},\quad k=1,\dots,K-1.
\end{equation}
We enforce the discrete convexity (increasing-slope) condition together with the monotonicity constraints:
\begin{equation}
\begin{aligned}
&s_{k} \le s_{k+1},\quad k=1,\dots,K-2,\\
&d^C_{1}<d^C_{2}<\cdots <d^C_{K},\\
&d^L_{1}\le d^L_{2}\le \cdots \le d^L_{K},
\end{aligned}
\label{eq.discrete_convex_monotonic_contraints}
\end{equation}
to ensure the discrete analogue of convexity (corresponding to nonnegative second differences), which effectively suppresses local extrema and oscillatory (“zig-zag”) behavior, while still permitting smooth, gradual variations in slope.

Finally, we pose anchor selection as a combinatorial problem:
\begin{equation}
\max_{\mathcal{S}\subseteq \tilde{\mathcal{A}}}\  |\mathcal{S}|, 
\quad \text{s.t. }\text{\eqref{eq.discrete_convex_monotonic_contraints} hold for }\mathcal{S}.
\label{eq.max_card}
\end{equation}
Problem \eqref{eq.max_card} seeks the longest monotone, convex subsequence, which represents the longest sequence with nondecreasing $d^C$, nondecreasing $d^L$, and nondecreasing secant slopes. 
In the implementation, we incorporate quadratic dynamic programming to identify the longest sequence, which is robust to outliers and promotes near-linearity while preserving coverage over $d^C$. The pseudo code is presented in Algorithm \clb{1} in the supplementary material.

\subsection{Network Architecture of DST-Calib}
\label{sec.network_architecture}

\begin{figure}[t!]
    \centering
    \includegraphics[width=0.98\linewidth]{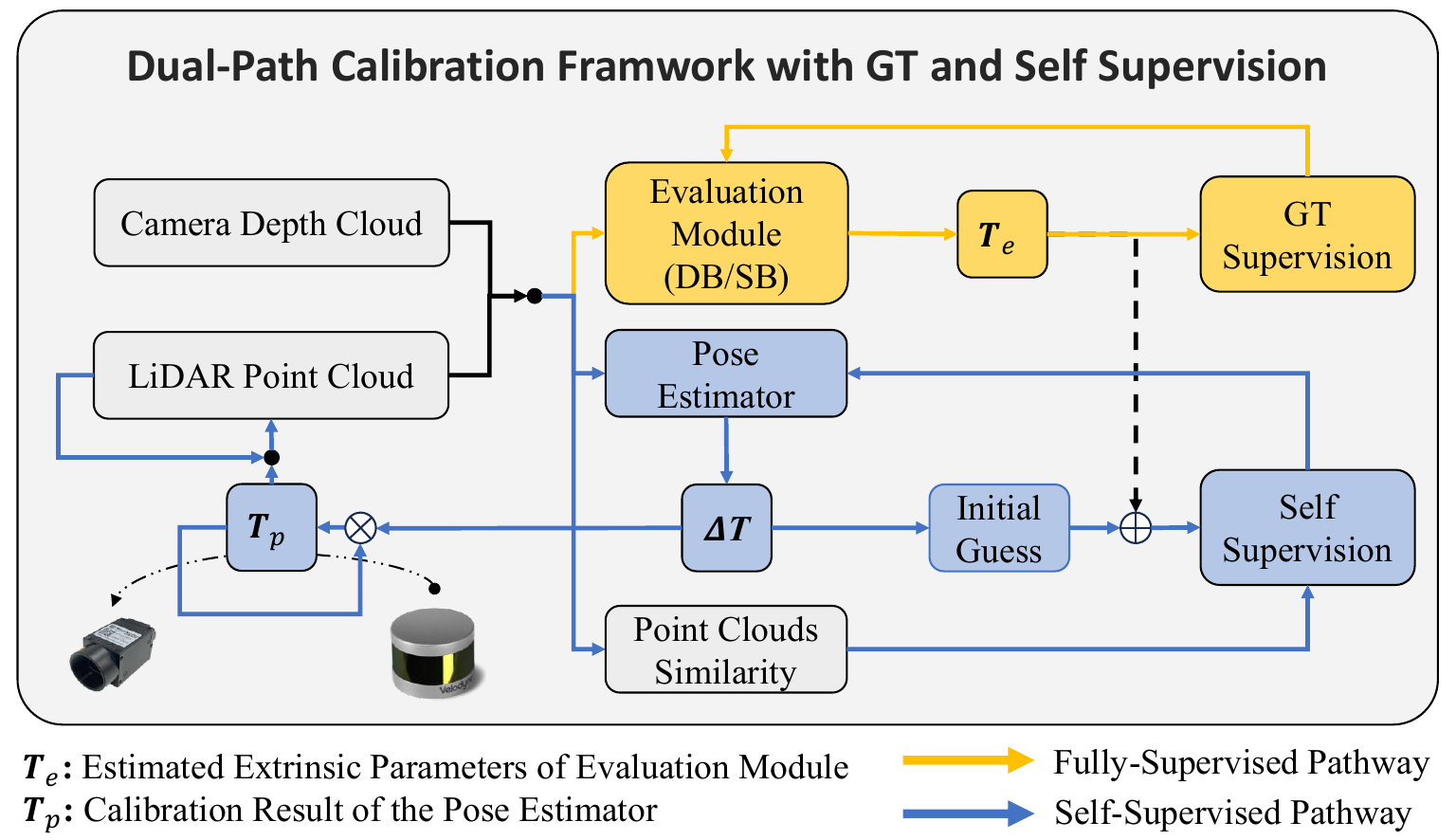}
    \caption{The proposed novel learning-based calibration framework with GT and self-supervision. This architecture has two optional pathways.}
    \label{fig.calibration_architecture}
\end{figure}

As depicted in Fig. \ref{fig.calibration_architecture}, DST-Calib adopts a novel framework, comprising two pathways: a fully-supervised pathway and a self-supervised pathway. In the fully-supervised pathway, DST-Calib employs an evaluation module to infer the similarity between the input LiDAR point cloud and camera depth cloud, along with the coarse estimation of the extrinsic parameters. The supervision signal of this pathway is produced from the calibration ground truth. In the self-supervised pathway, DST-Calib uses a pose estimator to infer 6-DoF extrinsic parameters, guided by point cloud similarity, an initial guess of the extrinsic parameters, and prior results (optional) from the evaluation network in the fully-supervised pathway. 
The two different pathways provide the following three options for specific usage, ensuring that it can achieve stable performance in various environmental and sensor setups:
\begin{itemize}
    \item{\textbf{Fully-Supervised Pathway Only}: When only activating the fully-supervised pathway, DST-Calib utilizes calibration ground-truth to supervise the training of the network. The extrinsic parameters are directly regressed using the evaluation module, and the similarity between the LiDAR point cloud and camera depth cloud can be estimated. In cases where the initial estimation of the extrinsic parameters is quite close to the ground-truth position, this method is efficient, highly accurate, and generalizable to various challenging conditions.}
    \item{\textbf{Self-Supervised Pathway Only}: When only activating the self-supervised pathway, DST-Calib utilizes the pose estimator to derive extrinsic parameters with the supervision from the point cloud similarity, the initial guess of the extrinsic parameters, and, if possible, the evaluated prior given by the evaluation module. This application form is especially suitable for the rotation-only calibration task.}
    \item{\textbf{Both Pathway Activated}: When both pathways are activated, DST-Calib can achieve robust calibration with high accuracy and generalization ability even under a very large mis-calibration range. If the training data for the evaluation module are available, this method is strongly recommended for promising performance.}
\end{itemize}
\begin{figure}[t!]
    \centering
    \includegraphics[width=0.986\linewidth]{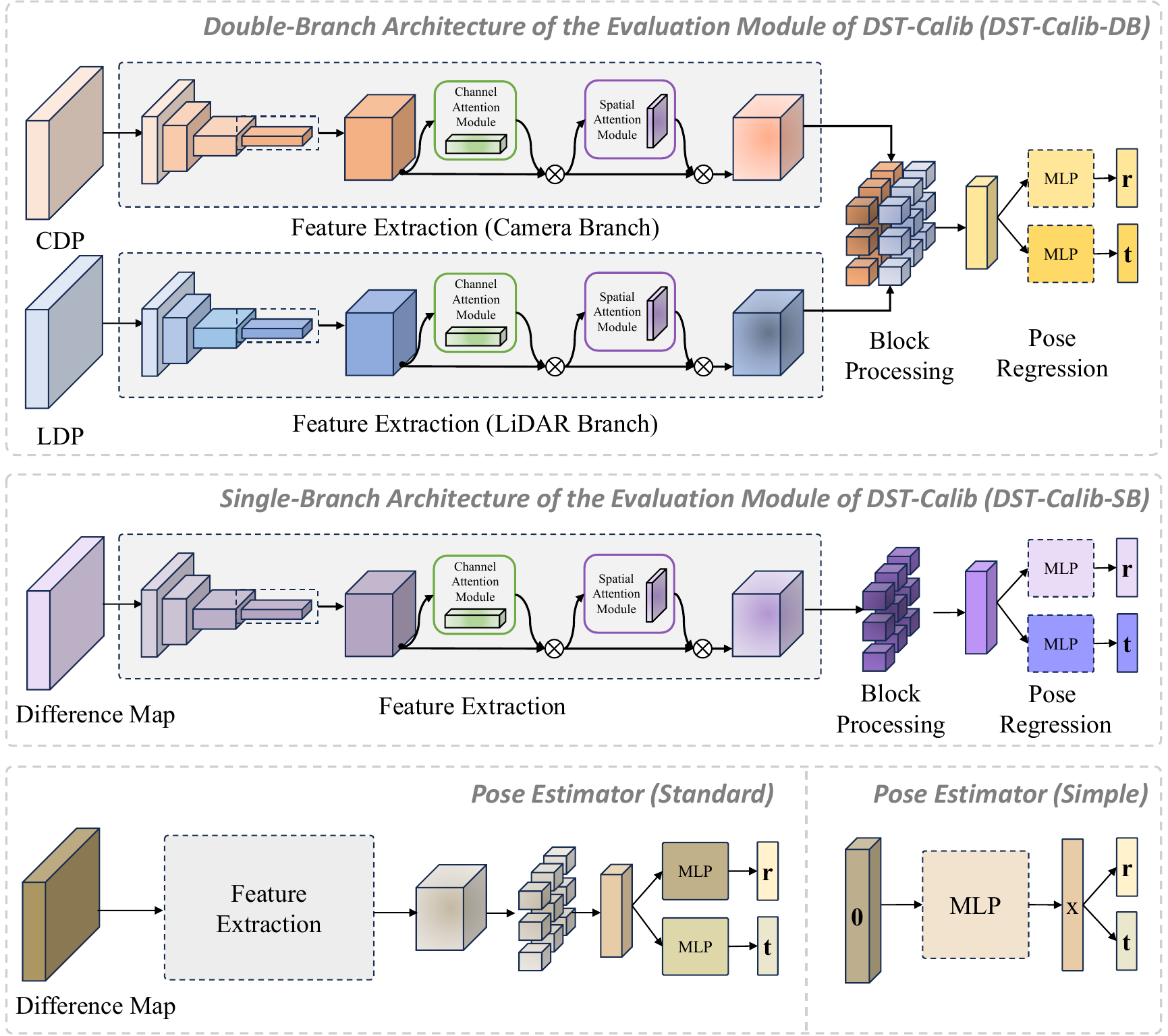}
    \caption{The main components of the proposed DST-Calib: DB, SB, and PE.}
    \label{fig.ComponentsOfZVNet}
\end{figure}

As illustrated in Fig. \ref{fig.ComponentsOfZVNet}, the main components of DST-Calib include the evaluation module and the pose estimator (PE). For the evaluation module, two different architectures, double-branch (DB) and single-branch (SB), are designed to explore the fundamental difference between the classical double-branch structure and our proposed single-branch architecture. Specifically, DB incorporates two separate branches for LiDAR and camera input, whereas SB has only one branch that receives the constructed difference map derived from the raw LiDAR point cloud and camera depth cloud. Both the DB and SB architectures incorporate CBAM modules to enhance feature extraction and cross-modal correlation. Additionally, block processing is added to further improve feature correlation (by combining different regions of LiDAR and camera inputs). In the following subsection, we will introduce the details of the most important parts: difference map generation, feature extraction and correlation, pose regression, and the pose estimator.

\subsubsection{Difference Map Construction}
\label{sec.diff_map_construction}

The primary goal of difference map construction is to effectively associate data from LiDAR and cameras, retaining the necessary information while demonstrating the differences. There are many potential ways to construct a difference map. For the calibration task, it is important to include depth information and the depth difference between the LiDAR point clouds and camera depth clouds. In this article, we propose a concise yet effective difference map generation strategy to fuse LiDAR and camera depth projections. Our approach preserves the original depth measurements while amplifying the discrepancies between the two modalities, thereby enhancing disparity cues. The procedure is detailed as follows.

Given the LiDAR depth projection image ${^L_D}{\boldsymbol{I}}$ and its corresponding rectified camera depth map ${^C_D}{\boldsymbol{I}}$ (two single-channel depth maps of size $H \times W$), we aim to construct a three-channel difference image $\boldsymbol{D}$ based on a threshold $e_{tar} > 0 \in \mathbb{R}$ ($e_{tar}$ is the target translation calibration error that satisfy the minimum requirement). Specifically, we define the pixel-wise difference as follows:
\begin{equation}
\Delta(u, v) = {^L_D}{\boldsymbol{I}}(u, v) - {^C_D}{\boldsymbol{I}}(u, v), 
u = 1, \dots, H;\, v = 1, \dots, W.
\end{equation}
Then, the difference map construction function is defined as follows:
\begin{equation}
\boldsymbol{D}(u, v) = \bigl( {^L_D}{\boldsymbol{I}}(u, v),\; [\Delta(u, v)]_+^{e_{tar}},\; [\Delta(u, v)]_-^{e_{tar}} \bigr),
\end{equation}
where
$$
[\Delta]_+^{e_{tar}} = 
\begin{cases}
\Delta, & |\Delta| > {e_{tar}}, \\
0,      & \text{otherwise},
\end{cases}
\quad
[\Delta]_-^{e_{tar}} = 
\begin{cases}
\Delta, & |\Delta| \le {e_{tar}}, \\
0,      & \text{otherwise}.
\end{cases}
$$
This obtained difference map $\boldsymbol{D}$ contains the necessary information for calibration, including the LiDAR depth, the difference of the LiDAR depth and camera estimated depth, and the depth projected coordinates.

\subsubsection{Feature Extraction} 
\label{sec.scene_discriminator}

We employ ResNet as the backbone for feature extraction, with both the double-branch (DB) and single-branch (SB) architectures adopting the same extraction pipeline. The input data are first transformed into a feature map $\boldsymbol{F_0}$, which is then processed by a CBAM module to generate an enhanced feature map $\boldsymbol{F}$ with spatial and channel attention.  

In the DB architecture, two independent branches are required to separately extract feature maps from the CDP and LDP inputs. By contrast, the SB architecture requires only a single extraction branch, since the LiDAR and camera data are explicitly associated through the constructed difference map. While the classical DB architecture extracts features from two distinct modalities and relies on an uncertain feature correlation process lacking theoretical justification, the SB design explicitly correlates the LiDAR point cloud and the camera depth cloud within a unified branch. This not only reduces the number of model parameters but also improves interpretability.

\subsubsection{Block Processing and Pose Regression}
\label{sec.method_dualpath_matching}

Unlike most previous methods that directly concatenate feature maps, we adopt a block-based processing strategy to enhance the regional relationships within the extracted feature map. Specifically, after feature extraction, the feature map $\boldsymbol{F}$ is divided into $n \times n$ grid blocks $\boldsymbol{F_i}$. Each block is processed by a convolutional module and compressed into a higher-level feature vector $\boldsymbol{B_i}$. The resulting block features are concatenated and unfolded into a linear feature vector $\boldsymbol{F}_{p}$, which is subsequently passed through a series of fully connected layers for global feature aggregation.  
Finally, we decouple the output streams for rotation and translation to account for potential modality-specific differences. Unless otherwise specified, the network's output is defined as a pose vector $\xi = \begin{bmatrix}
\boldsymbol{r}^\top & \boldsymbol{t}^\top
\end{bmatrix} \in \mathbb{R}^{1 \times 6}$, where $\boldsymbol{r}$ is the estimated rotation vector and $\boldsymbol{t}$ is the estimated translation vector. $\boldsymbol{T}$ is the corresponding extrinsic transformation matrix.

\subsubsection{Pose Estimator}
\label{sec.method_pe}
As illustrated in Fig. \ref{fig.ComponentsOfZVNet}, the pose estimator in DST-Calib is a self-supervised module that leverages point cloud similarity to guide the estimation of extrinsic parameters. We design two possible realizations of this module. The first adopts a standard architecture similar to that of the evaluation module, enabling feature-based regression of the 6-DoF pose. The second employs a lightweight structure consisting of a multilayer perceptron (MLP) with a constant zero vector input, which functions as an automatic optimizer. In this case, the extrinsic transformation is iteratively updated using the feedback from cloud similarity and the calibration errors provided by the evaluation module. 
Using this pose estimator, DST-Calib can perform calibration in a fully self-supervised manner, without any ground-truth labels or retraining. 
Additionally, the evaluation module focuses solely on learning point cloud similarity, which is configuration-agnostic, and provides robust feedback to guide the self-supervised pose estimator across diverse sensor setups.

\subsection{Supervision Method}
\label{sec.supervision}

Unlike other DNN-based LCEC methods, the extrinsic ground truth $^{C}_{L}\boldsymbol{T}$ of the training dataset is not necessary for the supervision. Otherwise, we utilize the similarity between the LiDAR point cloud and the estimated camera depth cloud to supervise the network. However, if the extrinsic ground-truth is provided, we can also use it to enhance calibration performance by pre-training the evaluation module and using the pre-trained model to reinforce the similarity calculation, thereby improving overall calibration accuracy. In this section, we will provide details of the supervision method, including both full supervision for the evaluation module and self-supervision for the pose estimator.

\subsubsection{Supervision for the Evaluation Module}
For the evaluation module, we utilize the ground-truth extrinsic parameters from LiDAR to camera to supervise the training process. The loss function contains three components: rotation error, translation error, and point cloud distance error.

The rotation error is defined as follows:
\begin{equation}
\mathcal{L}_{r_{gt}} = \left\|\boldsymbol{R}_{cam}({{\boldsymbol{R}}}\ {\boldsymbol{R}_{lidar}})^{-1} - \boldsymbol{I}\right\|_{1,1}.
\label{eq.fully_supervise_rot}
\end{equation}

The translation error is defined as follows:
\begin{equation}
\mathcal{L}_{t_{gt}} = \left\|\boldsymbol{t}_{cam} - ({\boldsymbol{t}}_{lidar} + {{\boldsymbol{t}}})\right\|_{2}.
\label{eq.fully_supervise_trans}
\end{equation}

And the point cloud distance loss is defined as follows:
\begin{equation}
\begin{aligned}
\mathcal{L}_{cloud} = \sum_{\boldsymbol{p} \in \mathcal{P}}\bigg\| {{\boldsymbol{R}}}&({\boldsymbol{R}_{lidar}}\boldsymbol{p} + {\boldsymbol{t}_{lidar}}) + {{\boldsymbol{t}}} \\
-&({\boldsymbol{R}_{cam}}\boldsymbol{p} + {\boldsymbol{t}_{cam}})\bigg\|_{2},
\end{aligned}
\label{eq.fully_supervise_dist}
\end{equation}
which indicates the distance between the calibrated point cloud and the point cloud transformed using the ground truth. $\boldsymbol{p} = [x, y , z]^\top$ is a point in the input LiDAR point cloud $\mathcal{P}$.

Finally, the total loss function is the sum of the three components:
\begin{equation}
\mathcal{L}_{eva} = \mathcal{L}_{r_{gt}} + \mathcal{L}_{t_{gt}} + \mathcal{L}_{cloud}.
\label{eq.fully_supervise_all}
\end{equation}
With ground truth supervision, the evaluation module can learn the uncertainty component of the similarity between the LiDAR point cloud and the estimated camera depth cloud. This helps it perform well when calibrating the extrinsic parameters within a relatively small range and better guides the pose estimator in a much larger mis-calibration range.
\subsubsection{Supervision for the Pose Estimator}
\label{sec.supervision_pose_est}
There are three components in the supervision for the pose estimator. The first component is $\mathcal{L}_{t_{ini}}$, which is the in the same manner with $\mathcal{L}_{t_{gt}}$. However, the ${\boldsymbol{t}_{gt}}$ is replaced by the initial guess of the extrinsic parameters.

The second component is the chamfer distance between the transformed LiDAR point cloud and the estimated camera depth cloud, which is defined as follows:
\begin{equation}
\mathcal{L}_{CD} = \frac{\alpha}{|\hat{\mathcal{P}}|} \sum_{\hat{\boldsymbol{p}} \in \hat{\mathcal{P}}} \min_{\boldsymbol{q} \in \mathcal{Q}} \|\hat{\boldsymbol{p}} - \boldsymbol{q}\|_2^2
\;+\;
\frac{\beta}{|\mathcal{Q}|} \sum_{\boldsymbol{q} \in \mathcal{Q}} \min_{\hat{\boldsymbol{p}} \in \hat{\mathcal{P}}} \|\boldsymbol{q} - \hat{\boldsymbol{p}}\|_2^2,
\end{equation}
where $\hat{\mathcal{P}}$ is the transformed LiDAR point cloud using the estimated extrinsic matrix and $\mathcal{Q}$ is the camera depth cloud. The $\alpha$ and $\beta$ are the weights for the two components of the Chamfer distance and are set to 0.5 by default.

The third component is the extrinsic parameter evaluation score of the output extrinsic matrix $\boldsymbol{T}$ produced by the pose estimator, using the regressed pose $\boldsymbol{T}_{eva}$ from the evaluation module as a reference:
\begin{equation}
\mathcal{L}_{eva}' =  a\left\|\boldsymbol{e}_{eva} -{\boldsymbol{e}}\right\|_{2} + \left\|\boldsymbol{t}_{eva} - {\boldsymbol{t}} \right\|_2,
\end{equation}
where ${\boldsymbol{e}}$ is the corresponding Euler angle of the rotation component ${\boldsymbol{R}}$ of ${\boldsymbol{T}}$. $a$ is a scale parameter that balances the rotation and translation error. Considering that an extrinsic calibration error within $1^\circ$ in rotation and $0.1$m in translation is regarded as a good result, we set the scale parameter $a$ to $a = 0.1/1 = 0.1$.
The final loss $\mathcal{L}_{pe}$ is also the sum of these three components:
\begin{equation}
\mathcal{L}_{pe} = \mathcal{L}_{t_{ini}} + \mathcal{L}_{CD} + \mathcal{L}_{eva}'.
\label{eq.loss_self_supervise_all}
\end{equation}

If the initial guess of the extrinsic parameters is unavailable, $\mathcal{L}_{t_{ini}}$ should equal 0. And if the evaluation module cannot be pretrained in the application, $\mathcal{L}_{eva}'$ should also equal 0. So, this ensures that even if prior calibration results are not provided, and there is no condition to pre-train an evaluation module, DST-Calib can still achieve self-supervised learning with supervision from the point cloud chamfer distance.

\subsection{Multi-Frame Optimization}
\label{sec.multi_frame_opt}
Apart from directly inferring the extrinsic parameters from each frame, we also developed a multi-frame optimization method to jointly refine the calibration results using the constraints from data collected from multiple scenes under the same sensor setup. This multi-frame optimization can leverage multiple frames of LiDAR point clouds and camera images to jointly optimize the extrinsic calibration result. 
Let $\mathcal{T} = \{\boldsymbol{T}_1, \boldsymbol{T}_2, \ldots, \boldsymbol{T}_n\}$ be a series of extrinsic matrices that were obtained from the multi-frame output of DST-Calib, and let $h: SE(3) \rightarrow \mathbb{R}$ be a scoring function that assigns a quality score to each extrinsic matrix. Given a selection ratio $x \in (0, 1]$, we aim to compute the average of the top $x$ percent extrinsic parameters based on their scores.

In multi-frame optimization, each $\boldsymbol{T}_i$ is evaluated using a scoring function to measure its reliability. In the fully-supervised pathway, the scoring function is defined as follows:
\begin{equation}
\begin{aligned}
s_i & = h_{f}(\boldsymbol{T}_i,\boldsymbol{T}_i') \\
&= \exp \bigg(- (a\left\|\boldsymbol{e}_i'- \boldsymbol{e}\right\|_{2} + \left\|\boldsymbol{t}_{i}' - \boldsymbol{t}_{i} \right\|_2)\bigg), \quad i = 1, 2, \ldots, n,
\end{aligned}
\end{equation}
where $\boldsymbol{e}_i$ is the corresponded Euler angle of the rotation component $\boldsymbol{R}_i$ of $\boldsymbol{T}_i$, and $\boldsymbol{t}_i$ is the translation components of $\boldsymbol{T}_i$, respectively. $\boldsymbol{T}_i'$ is the evaluation result using the evaluation module with the input data at the position of $\boldsymbol{T}_i$.

In the self-supervised pathway, the scoring function is constructed using the Chamfer distance between the transformed LiDAR point cloud and the camera depth cloud, which is defined as follows:
\begin{equation}
\begin{aligned}
s_i & = h_s(\boldsymbol{T}_i) = \exp\big(-\mathcal{L}_{CD}(\hat{\mathcal{P}}_i,\mathcal{Q}_i)\big), \quad i = 1, 2, \ldots, n,\\
\end{aligned}
\end{equation}
where $\hat{\mathcal{P}}_i$ is the transformed LiDAR point cloud using the extrinsic matrix $\boldsymbol{T}_i$ and $\mathcal{Q}_i$ is the corresponded camera depth cloud in the $i$-th frame.

After calculating the score of each estimated extrinsic matrix, the initial matrices $\mathcal{T}$ are then ranked in descending order of their scores. Let $\pi: \{1, 2, \ldots, n\} \rightarrow \{1, 2, \ldots, n\}$ be a permutation such that $s_{\pi(1)} \geq s_{\pi(2)} \geq \cdots \geq s_{\pi(n)}$, the number of extrinsic matrices to select is determined by:
\begin{equation}
k = \left\lceil {x \cdot n}\right\rceil,
\end{equation}
where $\lceil \cdot \rceil$ denotes the ceiling function, ensuring at least one vector is selected when $x > 0$. The selected subset $\mathcal{T}_{\text{top}}$ contains the top $k$ vectors $\mathcal{T}_{\text{top}} = \left\{ \boldsymbol{T}_{\pi(1)}, \boldsymbol{T}_{\pi(2)}, \ldots, \boldsymbol{T}_{\pi(k)} \right\}$. Then, the average of the selected vectors can be computed using either uniform weighting or score-based weighting. In score-based weighting, matrices with higher scores contribute more to the average. The weight $w_j$:
\begin{equation}
w_j = \frac{s_{\pi(j)}}{\sum_{i=1}^{k} s_{\pi(i)}}, \quad j = 1, 2, \ldots, k
\end{equation}
for each selected matrix is proportional to its score. Using the selected extrinsic matrices in $\mathcal{T}_{\text{top}}$, we calculate the joint average of the translation and rotation components, $\boldsymbol{R}^*$ and $\boldsymbol{t}^*$, to obtain the final optimized extrinsic parameters $\boldsymbol{T}^*$.

\section{Experiment}
\label{sec.experiment}

In this section, we will detail and analyze the extensive experimental evaluations we performed to validate our novel
DST-Calib. First, we present the datasets we used to train and evaluate our approach, followed by the evaluation metrics and implementation details. Second, we report the generalization degradation caused by the single-sided data augmentation of the previous methods and analyze how the training strategy impacts the generalization. We then report qualitative and quantitative results compared with other SoTA target-free LCEC approaches and evaluate the performance of the proposed depth refinement method, DAR. Finally, we demonstrate the cross-domain generalization results of DST-Calib and conduct extensive ablation studies to validate each component of our methods.

\subsection{Datasets}
\label{sec.exp_datasets}
\begin{figure}[t!]
    \centering
    \includegraphics[width=0.98\linewidth]{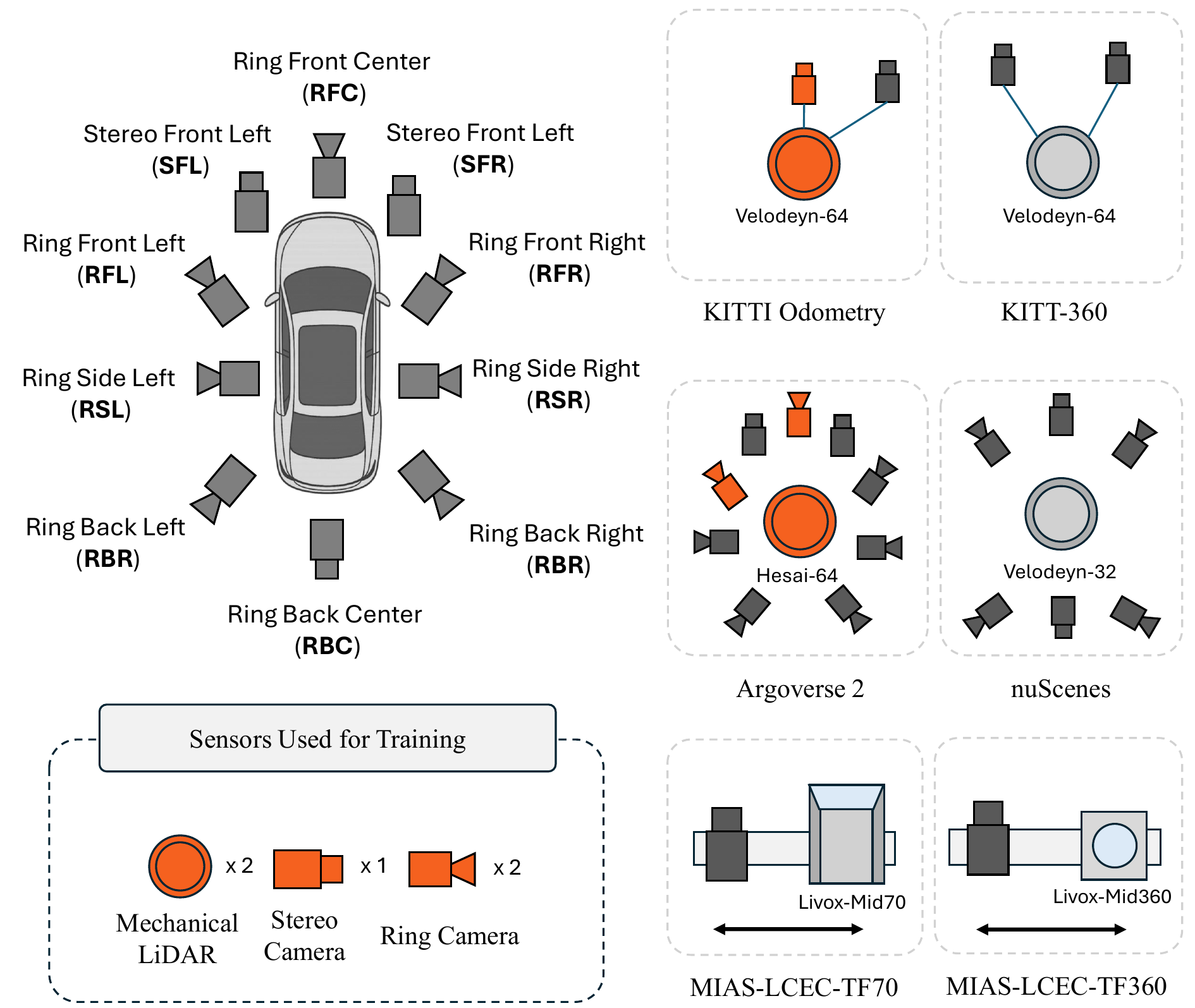}
    \caption{The different sensor configurations of the training and testing datasets. The orange-colored sensors are used for training, and the others are used for testing. Notably, in comparison with DNN-based methods, only the left camera of the KITTI 08 sequence is utilized for training to ensure a fair comparison.}
    \label{fig.sensors_configuration_define}
\end{figure}

We have conducted extensive experiments on the public dataset KITTI Odometry \cite{geiger2012we}, KITTI-360 \cite{liao2022kitti}, MIAS-LCEC \cite{zhiwei2024lcec} (including TF70 and TF360), nuScenes \cite{caesar2020nuscenes}, Argoverse2 \cite{wilson2023argoverse}, and our newly recorded dataset LCScenes. These real-world datasets are recorded in different countries, with different sensor setups, and include different types of calibration scenarios. Table \ref{tab.datasets_statics} summarizes the main characteristics of the datasets we used for training and evaluation. Fig. \ref{fig.sensors_configuration_define} illustrates the details of the training and testing sensor configuration. 

\begin{figure}[t!]
    \centering
    \includegraphics[width=0.987\linewidth]{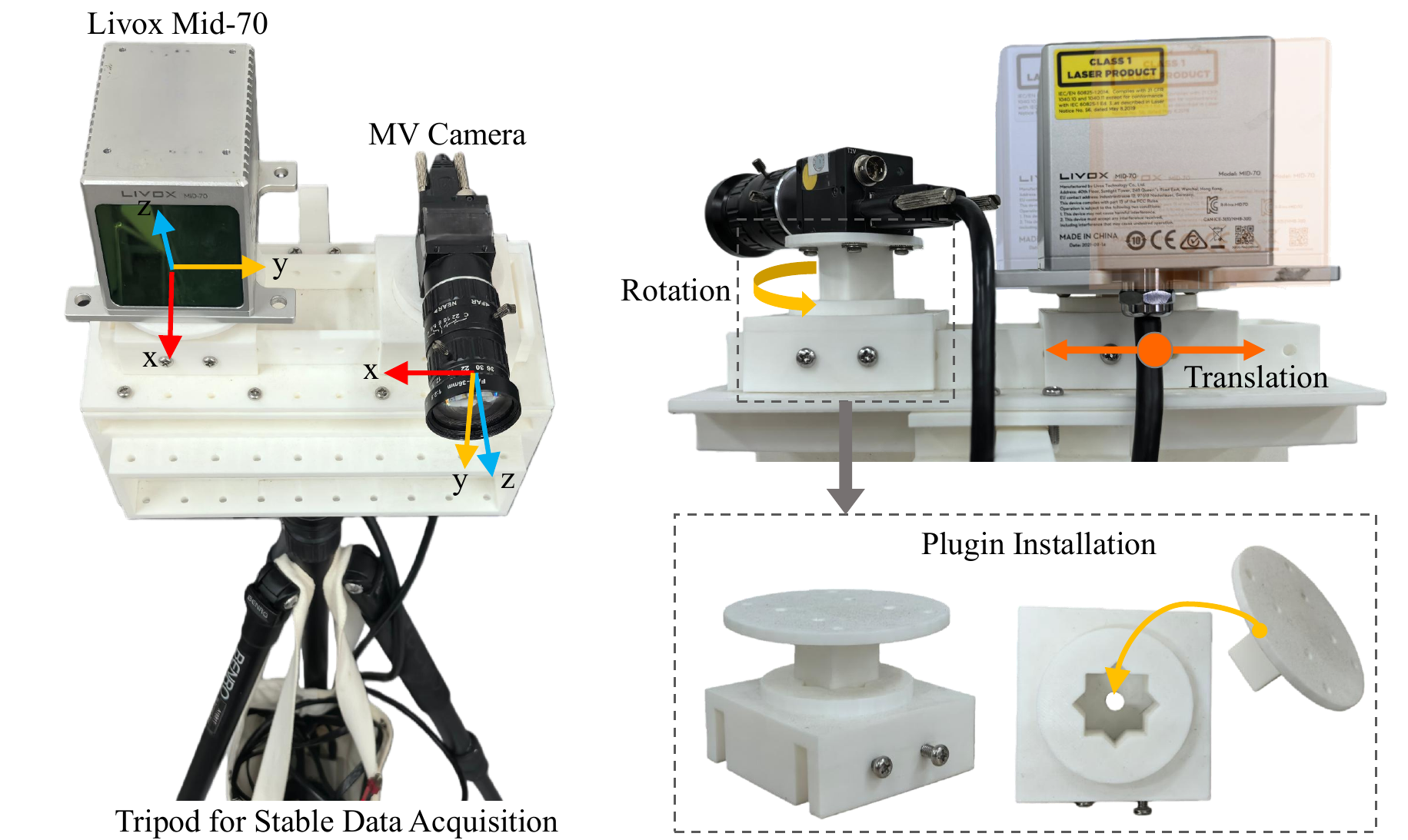}
    \caption{The data acquisition platform of LCScenes. A solid-state Livox LiDAR and one MindVision camera are utilized for capturing dense 4D point clouds and RGB images.}
    \label{fig.data_acquisition}
\end{figure}

LCScenes is recorded using our customized platform, shown in Fig. \ref{fig.data_acquisition}. The LiDAR and camera are installed on an adjustable platform placed on a tripod. It contains extensive pairs of 4D point clouds and 2D camera images with different extrinsic parameters, captured in different indoor and outdoor scenarios. This dataset is divided into five sequences, each recorded along a specific route. The LiDAR sensor is Livox-Mid70. The resolutions of the camera sensors are 1200 × 800 and 2400 × 1200.

\begin{table}[t!]
\caption{
Sensor setups in the evaluation datasets.}
\centering
\fontsize{6.7}{10}\selectfont
\begin{tabular}{l|c@{\hspace{0.15cm}}c@{\hspace{0.15cm}}c|c}
\toprule
{Dataset} & LiDAR &\makecell{Cameras} &\makecell{Camera \\Resolution} & \makecell{Sensors Included \\ in Training}\\
\hline
\hline
KITTI Odometry&  HDL-64E &2 &\makecell{1226 × 370 /\\1241 × 376} &\makecell{1 × LiDAR \\ 1 × camera}   \\
\hline
KITTI-360&  HDL-64E &2 &\makecell{1408 × 376} &  \ding{55} \\
\hline
Argoverse2 &2 × VLP-32 &9 &\makecell{2048 × 1550 /\\ 1550 × 2048} &\makecell{1 × LiDAR \\ 2 × camera}\\
\hline
MIAS-LCEC-TF70 &Livox-Mid70 &1 & 1200 × 800  &\ding{55} \\
\hline
MIAS-LCEC-TF360 &Livox-Mid360 &1&1200 × 800 &\ding{55}\\
\hline
nuScenes &HDL-32E &6&\makecell{1600 × 900 /\\1600 × 1200}&\ding{55}\\
\hline
LCScenes &Livox-Mid70 &2&\makecell{1200 × 800 /\\2400 × 1200}&\ding{55}\\
\bottomrule
\end{tabular}
\label{tab.datasets_statics}
\end{table}
\begin{table*}[t!]
\caption{Quantitative results of SoTA DNN-based approaches on the left and right camera of KITTI Odometry. The best results are shown in bold type, and the second-best ones are underlined.}
\centering
\fontsize{6.4}{10}\selectfont
\begin{tabular}{l|c|c@{\hspace{0.15cm}}c|c@{\hspace{0.15cm}}c@{\hspace{0.15cm}}c|c@{\hspace{0.15cm}}c@{\hspace{0.15cm}}c|c@{\hspace{0.15cm}}c|c@{\hspace{0.15cm}}c@{\hspace{0.15cm}}c|c@{\hspace{0.15cm}}c@{\hspace{0.15cm}}c}
\toprule
\multirow{3}*{Approach}& \multirow{3}*{Initial Range}&\multicolumn{8}{c|}{Left Camera} &\multicolumn{8}{c}{Right Camera}\\
\cline{3-18}
&&\multicolumn{2}{c|}{Magnitude}
&\multicolumn{3}{c|}{Rotation Error ($^\circ$)} &\multicolumn{3}{c|}{Translation Error (m)} 
&\multicolumn{2}{c|}{Magnitude}
&\multicolumn{3}{c|}{Rotation Error ($^\circ$)} &\multicolumn{3}{c}{Translation Error (m)}\\

&& $e_r$ ($^\circ$) & $e_t$ (m) & Yaw & Pitch & Roll  & {X} &  {Y} &  {Z}   & $e_r$ ($^\circ$) & {$e_t$ (m)} & Yaw & Pitch & Roll   &  {X} &  {Y} &  {Z}\\
\hline
\hline
CalibDNN &$[\pm10^\circ, \pm 1.0\text{m}]$ &1.172&0.098&0.402& 0.998&0.180&0.072&0.025&0.045 
&1.238&0.405&0.480 &1.010  &0.195&0.396 &0.026  &0.050 \\
CalibDNN (No RGB)  &$[\pm10^\circ , \pm 1.0\text{m}]$ &1.169&0.096&0.405&0.994&0.171&0.070 &0.024 &0.045 
&1.218&0.407&0.473&0.999 &0.184&0.397 & 0.026 &0.050 \\
CalibDNN  &$[\pm5^\circ , \pm 0.5\text{m}]$ &\textbf{0.585}&0.059&\textbf{0.201}& 0.493&0.106&0.042&\textbf{0.015}&\textbf{0.027} 
&0.692&0.328&\textbf{0.330}&0.512&0.141&0.323&0.017 &\underline{0.033} \\
CalibDNN (No RGB)  &$[\pm5^\circ , \pm 0.5\text{m}]$ &\underline{0.605}&\underline{0.058}&\underline{0.202}&0.506&0.128&0.041&\underline{0.016} &\textbf{0.027}
&\underline{0.685}&0.317&\underline{0.331} &0.493 &0.158&0.312 &0.018 &\textbf{0.032} \\
\hline
CalibDepth  &$[\pm10^\circ , \pm 1.0\text{m}]$ &0.996&0.075&0.332& 0.848 &0.141&0.046 &0.023 &0.038
&1.140&0.406&0.469 &0.915&0.166&0.392 &0.029 &0.049 \\
CalibDepth (No RGB)  &$[\pm10^\circ , \pm 1.0\text{m}]$ &1.106&0.073&0.332 &0.971 &0.134&0.045 &0.023&0.038
&1.181&0.409&0.466 &0.966  &0.161&0.396 &0.027 &0.048 \\
CalibDepth  &$[\pm5^\circ , \pm 0.5\text{m}]$ &0.638&\textbf{0.052}&0.268 &\underline{0.498} &0.137&\underline{0.025} &0.017 &0.036 
&0.904&\textbf{0.303}&0.520 &0.507  &0.332&\textbf{0.290} &0.025 &0.046 \\
CalibDepth (No RGB)  &$[\pm5^\circ , \pm 0.5\text{m}]$ &0.630&\textbf{0.052}&0.255 &0.501&0.133&\textbf{0.024} &0.018  &0.036
&0.913&0.305&0.526 &0.505 &0.333&\underline{0.293}  &0.024&0.045 \\
\hline
CalibNet &$[\pm10^\circ , \pm 1.0\text{m}]$ &1.109&0.167&0.333 &0.955 &0.206&0.115  &0.046 &0.086
&1.182&0.389&0.346 &1.018 &0.226&0.363 &0.048 &0.084 \\
CalibNet (No RGB)  &$[\pm10^\circ , \pm 1.0\text{m}]$ &1.251&0.164&0.575 &0.984 &0.190&0.117 &0.042 &0.079
&1.238&0.407&0.510 &0.993 &0.224&0.384 &0.050 &0.082 \\
CalibNet  &$[\pm5^\circ , \pm 0.5\text{m}]$ &0.666&0.067&0.323 &0.515&\underline{0.080}&0.046&0.017&0.034
&\textbf{0.665}&\underline{0.304}&0.356&\textbf{0.481}&\underline{0.103}&0.298&\underline{0.016}   &0.036 \\
CalibNet (No RGB)  &$[\pm5^\circ , \pm 0.5\text{m}]$ &0.650&0.068&0.336 &\textbf{0.486} &\textbf{0.070}&0.049 &\underline{0.016} &\underline{0.033} &0.702&0.309&0.411 &\underline{0.483} &\textbf{0.089}&0.303 &\textbf{0.015} &0.037 \\

\bottomrule
\end{tabular}
\label{tab.general_degrade}
\end{table*}

\subsection{Evaluation Metrics}
To comprehensively evaluate the performance of LCEC approaches, we follow the previous works \cite{lv2021lccnet,koide2023general} to use the magnitude $e_r$ of Euler angle error and the magnitude $e_t$ of the translation error, with the following expression:
\begin{equation}
\begin{aligned}
e_r &=  \left\|{\boldsymbol{e}^*} - {\boldsymbol{e}_{gt}}\right\|_2, \\
e_t &= \left\|{\boldsymbol{t}}^* - {\boldsymbol{t}}_{gt}\right\|_2,
\end{aligned}
\label{eq.exp_metrics}
\end{equation}
{to quantify the calibration errors. In (\ref{eq.exp_metrics}), ${\boldsymbol{e}^{*}}$ and ${\boldsymbol{e}_{gt}}$ represent the estimated and ground-truth Euler angle vectors, computed from the rotation matrices ${\boldsymbol{R}}^*$ and ${\boldsymbol{R}_{gt}}$, respectively. Similarly, $\boldsymbol{t}^*$ and $\boldsymbol{t}_{gt}$ denote the estimated and ground-truth translation vectors from LiDAR to camera, respectively.}

In monocular depth estimation, following previous works \cite{eigen2014depth, godard2019digging,feng2022advancing}, performance is evaluated using several standard metrics. Mean Absolute Error (MAE) and Root Mean Squared Error (RMSE) measure the absolute and squared differences between predicted and ground-truth depths, capturing overall accuracy and penalizing large deviations. Absolute Relative Error (Abs Rel) and Squared Relative Error (Sq Rel) assess the relative discrepancy between predictions and ground truth by normalizing the errors with respect to true depth values, making them sensitive to accuracy across different distance ranges. Additionally, the threshold accuracy metrics $\delta_1$, $\delta_2$, and $\delta_3$ compute the percentage of predicted depths that lie within progressively permissive multiplicative error bounds, reflecting the robustness of the estimation. 

\subsection{Training Details}
Unless otherwise specified, we train DST-Calib on a single NVIDIA RTX4090D GPU with a total batch size of 8, using the AdamW optimizer with a base learning rate of  $5*10^{-4}$ and a weight decay of $1*10^{-4}$. A OneCycle learning rate scheduler is used during training. For each instance of DST-Calib, the evaluation module is trained for 200 epochs using the fully supervised loss defined in \eqref{eq.fully_supervise_all}. For the pose estimator in the self-supervised pathway, we employ a simple architecture by default and directly apply it to multi-frame online calibration, assuming that the extrinsic parameters remain consistent across all frames in a sequence. The pose estimator is trained online, self-supervised, in parallel with the calibration process. All the frames in an online calibration data sequence are used for self-supervised learning of the pose estimator. If the number of frames is relatively small, we repeatedly sample the data to ensure at least 30 batches are processed during the learning process. 

Most previous DNN-based approaches train and test their model solely on the KITTI dataset \cite{lv2021lccnet,yuan2020rggnet,zhao2021calibdnn}. Notably, sequences in KITTI Odometry, aside from 00, were included in the training datasets for the DNN-based methods. To ensure a fair comparison, we reproduced calibration results for both the left and right cameras on sequence 00 when the authors provided their code; otherwise, we used the reported results for the left camera from their papers. For our DST-Calib, we train it on the KITTI Odometry 08 sequence and test it on the other sequences. Since most learning-based methods lack APIs for custom data, our comparison with these methods is limited to the KITTI Odometry 00 sequence. 

To further improve DST-Calib's generalization, we also train it using samples from KITTI Odometry and Argoverse2. These models trained on multiple datasets are used to validate the cross-domain generalization ability. In the training on multiple datasets, the left camera of the KITTI Odometry 08 sequence, the ring front center camera of Argoverse2 000\_00-000\_01, and the ring front left camera of Argoverse2 000\_00-000\_01 are utilized as the training samples. We used the pretrained weights on the KITTI Odometry 08 sequence as the initial model parameters. 

During the training and testing, unless otherwise specified, the random deviation range for both the LiDAR and camera side of the double-sided data augmentation is set to a total of $[\pm5^\circ,\pm0.5\text{m}]$ for all axes in translation and rotation. During the experiment and the reproduction of previous work, we found that the network is hard to converge when the uniform deviation along each axis is large. To make the training process easier to converge, a weight $[0.6,0.2,0.2]$ is set to each axis. For example, under the deviation $[\pm5^\circ,\pm0.5m]$, the exact deviation of each axis of rotation and translation is $[\pm3^\circ,\pm1^\circ,\pm1^\circ]$ and $[\pm0.3\text{m},\pm0.1\text{m},\pm0.1\text{m}]$, respectively. The virtual camera's projection size is set to (256, 512) and the focal length to 600 by default. Additionally, the block processor's block number is set to 5 by default.

\subsection{Inference}
\label{sec.exp_inference}

In the experiments, we evaluate five different types of DST-Calib, including DB, SB, SB*, PE, and PE+SB*, trained under different initial mis-calibration ranges. DB is the evaluation module in the fully-supervised pathway that adopts the double-branch architecture. SB is the evaluation module that adopts our proposed single-branch architecture. SB* is the single-branch evaluation module that utilizes the multi-frame optimization in the calibration. PE is the pose estimator in the self-supervised pathway. PE+SB* combines both fully-supervised and self-supervised ways. It utilizes the inference results of PE as the prior for SB to obtain a refined calibration result. Notably, both PE and PE+SB* adopt multi-frame optimization, where extrinsic parameters between LiDAR and camera are assumed to be the same across a sequence. For the testing of multi-frame optimization, we conduct 10 separate sequence calibrations (each with a distinct initial pose) and use the average value as the total calibration error to ensure a fair comparison with other instances.

\subsection{Generalization Degrade Caused by Unbalanced Single-Sided Data Augmentation}
\label{sec.exp_generalization_degrade}

\begin{figure}[t!]
    \centering
    \includegraphics[width=0.98\linewidth]{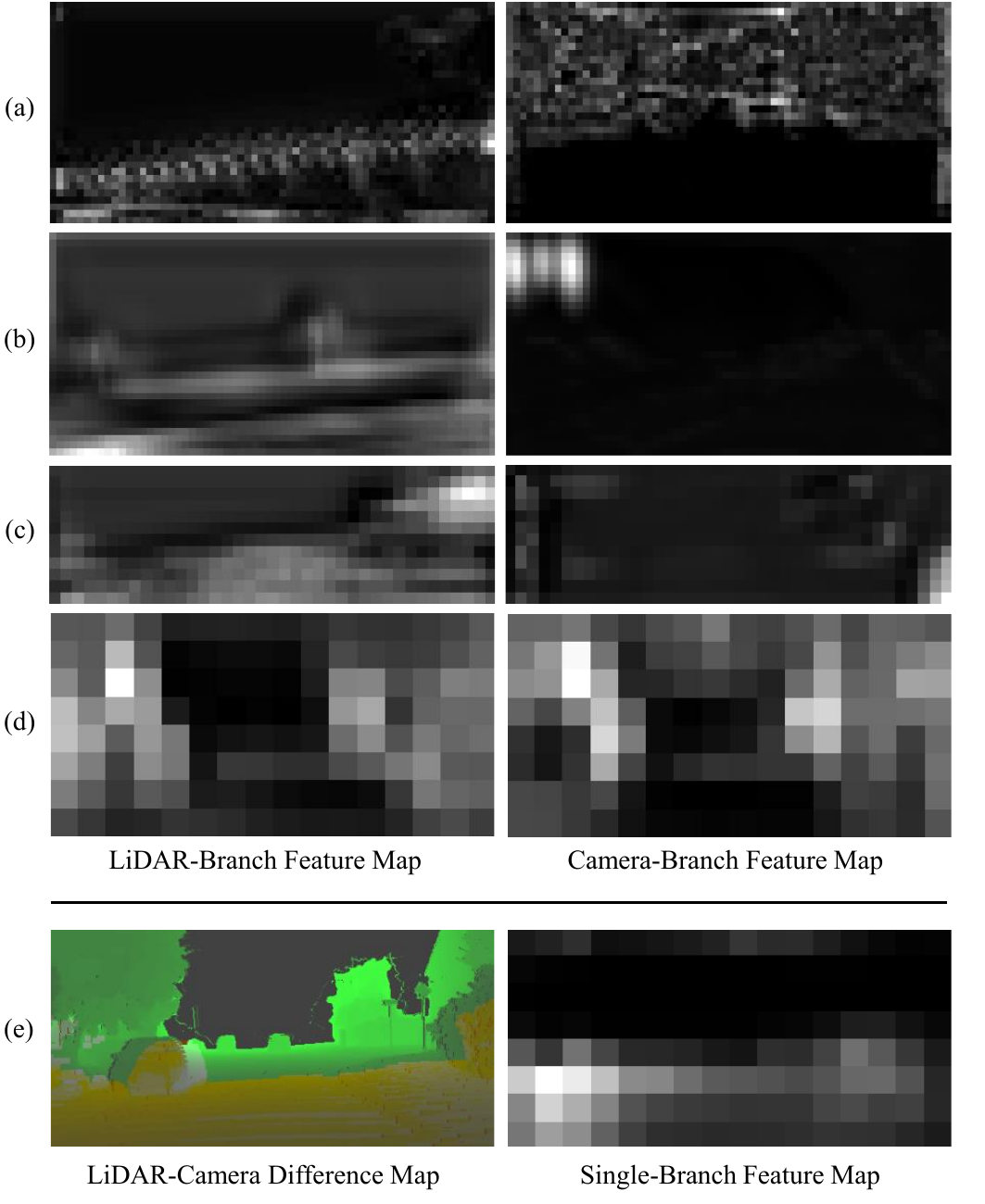}
    \caption{Visualization results of extracted feature maps: (a)-(d) The LiDAR-branch and camera-branch feature maps of CalibDepth, CalibNet, and CalibDNN; (e) the constructed LiDAR-camera difference map and its feature extraction result. Under the single-sided data augmentation, the camera-branch feature extraction results are sometimes almost empty and meaningless.}
    \label{fig.fea_map_cmp}
\end{figure}

In this section, we conduct extensive experiments to demonstrate the degradation in generation caused by the single-sided training data augmentation strategy (widely adopted in previous LCEC networks). Quantitative results are shown in Table \ref{tab.general_degrade}. Qualitative results are demonstrated in Fig. \ref{fig.fea_map_cmp}.

\subsubsection{Generalization Experiments using Different Sensor Configurations}

Extensive prior work trained and tested their networks on the left camera from KITTI sequences. However, the calibration task is a flexible mission that should adapt to various relative poses at different sensor locations. We test some typical DNN-based methods (including CalibDepth, CalibDNN, and CalibNet) on the KITTI right camera using their model trained on the left camera. As the experimental results shown in Table \ref{tab.general_degrade}, there is significant performance degradation when only changing the camera location within the same calibration scene. This shows that the trained model is not generalizable and only works for specific camera locations. 

\subsubsection{Degradation Phenomena in RGB Image Branch}
We have found that when the data augmentation range is unbalanced, the RGB branch degrades. As shown in Table \ref{tab.general_degrade}, even after removing the input RGB images from the camera branch of the model, it still achieves similar performance on the same sensor configuration. Additionally, we also trained the models using only the input from the LiDAR branch. They can also behave well on the training dataset. This further demonstrates that the single-sided data augmentation strategy and the double-branch model architecture are unable to effectively associate features between LiDAR point clouds and camera images. Under the many-to-one mapping of single-sided data augmentation, networks can directly ``guess'' relative poses from the shapes of LiDAR projections. Moreover, we visualize the feature extraction result of the camera input branch. As depicted in Fig. \ref{fig.fea_map_cmp}, the feature map after the feature extraction module of the camera image is almost empty and meaningless, which means that the input camera images have no contribution to the estimated extrinsic parameters of the networks. In contrast, with the double-sided data augmentation strategy, the feature maps produced by DST-Calib retain richer contextual information.

\begin{table}[t!]
\caption{
Quantitative comparisons of different training data augmentation ranges (using double-branch architecture).}
\centering
\fontsize{6.3}{10}\selectfont
\begin{tabular}{c|c|c|c@{\hspace{0.12cm}}c|c@{\hspace{0.12cm}}c}
\toprule
\multirow{2}*{$\boldsymbol{c}_{lidar}$} & \multirow{2}*{$\boldsymbol{c}_{cam}$} & \multirow{2}*{\makecell{Sensor \\Inputs}} & \multicolumn{2}{c|}{KITTI Left00} & \multicolumn{2}{c}{KITTI Right00}\\
& & &$e_r$ ($^\circ$) & $e_t$ (m)  & $e_r$ ($^\circ$) & $e_t$ (m)\\
\hline
\hline
\multirow{3}*{$[\pm5^\circ,\pm0.5 \text{m}]$} &\multirow{3}*{$[\pm0^\circ,\pm0.0 \text{m}]$}  
& Double &0.301	&0.067	&0.708	&0.192\\
& & No Camera &0.757	&0.091	&0.776	&0.229\\
& & No LiDAR &2.082	&0.221	&2.144	&0.237\\
\hline

\multirow{3}*{$[\pm5^\circ,\pm0.5 \text{m}]$} &\multirow{3}*{$[\pm1^\circ,\pm0.1 \text{m}]$}  
& Double &\textbf{0.168}	&\textbf{0.042}	&0.461	&0.137\\
& & No Camera &1.002	&0.169	&1.135	&0.294\\
& & No LiDAR &2.446	&0.228	&2.505	&0.257\\
\hline

\multirow{3}*{$[\pm5^\circ,\pm0.5 \text{m}]$} &\multirow{3}*{$[\pm2^\circ,\pm0.2 \text{m}]$}  
& Double &{0.191}	&0.046	&0.357	&0.100\\
& & No Camera &1.762	&0.154	&1.704	&0.211\\
& & No LiDAR &2.521	&0.250	&2.586	&0.246\\
\hline

\multirow{3}*{$[\pm5^\circ,\pm0.5 \text{m}]$} &\multirow{3}*{$[\pm3^\circ,\pm0.3 \text{m}]$}  
& Double &0.192	&\underline{0.043}	&0.253	&0.063\\
& & No Camera &1.409	&0.213	&1.376	&0.198\\
& & No LiDAR &2.105	&0.233	&2.056	&0.226\\
\hline

\multirow{3}*{$[\pm5^\circ,\pm0.5 \text{m}]$} &\multirow{3}*{$[\pm4^\circ,\pm0.4 \text{m}]$}  
& Double &{0.191}	&0.046	&\underline{0.232}	&\underline{0.058}\\
& & No Camera &2.162	&0.201	&2.190	&0.227\\
& & No LiDAR &2.678	&0.235	&2.666	&0.274\\
\hline

\multirow{3}*{$[\pm5^\circ,\pm0.5 \text{m}]$} &\multirow{3}*{$[\pm5^\circ,\pm0.5 \text{m}]$}  
& Double &\underline{0.186}	&0.045	&\textbf{0.218}	&\textbf{0.054}\\
& & No Camera &1.748	&0.249	&1.715	&0.276\\
& & No LiDAR &2.377	&0.241	&2.399	&0.248\\

\bottomrule
\end{tabular}
\label{tab.impact_of_different_gen_range}
\end{table}

\begin{table}[t!]
\caption{
Quantitative comparisons of two different architectures with or without the input of LiDAR and camera.}
\centering
\fontsize{6.5}{10}\selectfont
\begin{tabular}{c|c|c|c@{\hspace{0.18cm}}c|c@{\hspace{0.18cm}}c}
\toprule
\multirow{2}*{Data Source} &\multirow{2}*{Model} & \multirow{2}*{\makecell{Sensor \\Inputs}} & \multicolumn{2}{c|}{KITTI Left00} & \multicolumn{2}{c}{KITTI Right00}\\
& & &$e_r$ ($^\circ$) & $e_t$ (m)  & $e_r$ ($^\circ$) & $e_t$ (m)\\
\hline
\hline
\multirow{3}*{LDP-LDP} &\multirow{3}*{{DST-Calib (DB)}}  
& Double &\underline{0.186}	&\underline{0.045}	&\underline{0.218}	&\underline{0.054}\\
& & No Camera &1.748	&0.249	&1.715	&0.276\\
& & No LiDAR &2.377	&0.241	&2.399	&0.248\\
\hline

\multirow{3}*{LDP-CDP} &\multirow{3}*{{DST-Calib (DB)}}  
& Double &0.524	&0.101	&0.664	&0.147 \\
& & No Camera &1.554	&0.168	&1.590	&0.207 \\
& & No LiDAR &2.171	&0.207	&2.164	&0.207\\
\hline

\multirow{3}*{LDP-LDP} &\multirow{3}*{{DST-Calib (SB)}}  
& Double &\textbf{0.136}&\textbf{0.033} &\textbf{0.140}&\textbf{0.035}\\
& & No Camera &1.865&0.244 &1.885&0.242\\
& & No LiDAR &2.066&0.265 &1.978&0.264\\
\hline

\multirow{3}*{LDP-CDP} &\multirow{3}*{{DST-Calib (SB)}}  
& Double &0.366	&0.078 &0.505 &0.109\\
& & No Camera &2.115&0.174 &2.086&0.186\\
& & No LiDAR &1.900 &0.203 &1.837&0.197\\

\bottomrule
\end{tabular}
\label{tab.impact_of_diff_input_modal}
\end{table}

\begin{table*}[t!]
\caption{Quantitative comparisons with SoTA target-free LCEC approaches on the 00 sequence of KITTI Odometry. \dag: These methods did not release code, preventing the reproduction of results for both cameras.}
\centering
\fontsize{6.5}{10}\selectfont
\begin{tabular}{l|c|c@{\hspace{0.15cm}}c|c@{\hspace{0.15cm}}c@{\hspace{0.15cm}}c|c@{\hspace{0.15cm}}c@{\hspace{0.15cm}}c|c@{\hspace{0.15cm}}c|c@{\hspace{0.15cm}}c@{\hspace{0.15cm}}c|c@{\hspace{0.15cm}}c@{\hspace{0.15cm}}c}
\toprule
\multirow{3}*{Approach}& \multirow{3}*{Initial Range}&\multicolumn{8}{c|}{Left Camera} &\multicolumn{8}{c}{Right Camera}\\
\cline{3-18}
&&\multicolumn{2}{c|}{Magnitude}
&\multicolumn{3}{c|}{Rotation Error ($^\circ$)} &\multicolumn{3}{c|}{Translation Error (m)} 
&\multicolumn{2}{c|}{Magnitude}
&\multicolumn{3}{c|}{Rotation Error ($^\circ$)} &\multicolumn{3}{c}{Translation Error (m)}\\

&& $e_r$ ($^\circ$) & $e_t$ (m) & Yaw & Pitch & Roll  & {X} &  {Y} &  {Z}   & $e_r$ ($^\circ$) & {$e_t$ (m)} & Yaw & Pitch & Roll   &  {X} &  {Y} &  {Z}\\
\hline
\hline
CalibRCNN\textsuperscript{\dag} \cite{shi2020calibrcnn} &$[\pm10^\circ, \pm 0.25\text{m}]$ &0.805 &0.093	&0.446	&0.640	&0.199	&0.062	&0.043 &0.054 &- &-	&-	&-	&-	&-	&-	&-\\
RegNet\textsuperscript{\dag} \cite{schneider2017regnet} &$[\pm20^\circ, \pm 1.5\text{m}]$ &0.500 &0.108	&0.240	&0.250	&0.360	&0.070	&0.070	&0.040 &- &-&-	&-	&-	&-	&-	&-\\
LCCNet \cite{lv2021lccnet} &$[\pm10^\circ, \pm 1.0\text{m}]$ &1.418 &0.600  &0.455 &0.835 &0.768 &0.237 &0.333  &0.329 &1.556  &0.718 & 0.457 &1.023 &0.763 &0.416 &0.333 &0.337 \\
RGGNet \cite{yuan2020rggnet} &$[\pm20^\circ, \pm 0.3\text{m}]$ &1.290 &0.114	&0.640 &0.740&0.350 &0.081 &{0.028} &0.040 &3.870 &0.235 &1.480 &3.380 &0.510	&0.180 &{0.056} &0.061 \\
CalibDNN \cite{zhao2021calibdnn} &$[\pm10^\circ, \pm 1.0\text{m}]$ &1.172&0.098&0.402& 0.998&0.180&0.072&0.025&0.045 
&1.238&0.405&0.480 &1.010  &0.195&0.396 &0.026  &0.050 \\
CalibDepth \cite{zhu2023calibdepth}  &$[\pm10^\circ , \pm 1.0\text{m}]$ &0.996&0.075&0.332& 0.848 &0.141&0.046 &\textbf{0.023} &0.038
&1.140&0.406&0.469 &0.915&0.166&0.392 &0.029 &0.049 \\
CalibNet \cite{iyer2018calibnet} &$[\pm10^\circ , \pm 1.0\text{m}]$ &1.109&0.167&0.333 &0.955 &0.206&0.115  &0.046 &0.086
&1.182&0.389&0.346 &1.018 &0.226&0.363 &0.048 &0.084 \\
\hline
CRLF \cite{ma2021crlf} &- &0.629	&4.118	&\textbf{0.033}	&0.464	&0.416	&3.648	&1.483	&0.550 &0.633	&4.606	&\textbf{0.039}	&0.458	&0.424	&4.055	&1.636	&0.644\\
UMich \cite{pandey2015automatic} &- &4.161	&0.321	&0.113	&3.111	&2.138	&0.286	&0.077	&0.086
 &4.285&0.329	&\underline{0.108}	&3.277	&2.088	&0.290	&0.085	&0.090\\
HKU-Mars \cite{yuan2021pixel} &- &33.84	&6.355	&19.89	&18.71	&19.32	&3.353	&3.232	&2.419
&32.89	&4.913	&18.99	&15.77	&17.00	&2.917	&2.564	&1.646\\
DVL \cite{koide2023general}  &-&{122.1} &{5.129}	&{48.64}	&{87.29}	&{98.15}	&{2.832}	&{2.920}	&{1.881} &{120.5} &{4.357}	&{49.60}	&{87.99}	&{96.72}	&{2.086}	&{2.517}	&{1.816}\\
MIAS-LCEC \cite{zhiwei2024lcec}  &-&5.385	&1.014	&1.574	&4.029	&4.338	&0.724	&0.383	&0.343 &7.655	&1.342	&1.910	&5.666	&6.154	&0.843	&0.730	&0.358\\

EdO-LCEC \cite{huang2025environment}  &- &\underline{0.295}&0.082	&0.117	&0.176	&{0.150}	&0.051	&0.044	&\textbf{0.032} 
&\underline{0.336}	&{0.118}	&0.216	&\textbf{0.168}	&0.121	&\underline{0.083}	&0.067	&\underline{0.032} \\

\hline
DST-Calib (DB)  &2*$[\pm5^\circ , \pm 0.5\text{m}]$ &0.524	&0.101	&0.292	&0.308	&0.173	&0.077	&0.024	&0.044	&0.664	&0.147	&0.318	&0.440	&0.207	&0.129	&0.023	&0.045 \\
DST-Calib (DB)  &2*$[\pm10^\circ , \pm 1.0\text{m}]$ &1.132	&0.185	&0.634	&0.724	&0.284	&0.136	&0.042	&0.084	&1.196	&0.246	&0.683	&0.740	&0.320	&0.207	&0.046	&0.082 \\

DST-Calib (SB)  &2*$[\pm5^\circ , \pm 0.5\text{m}]$  &0.366	&\underline{0.078}	&0.195 &\underline{0.236} &0.092&0.054 &0.020 &0.038
&0.505	&\underline{0.109}	&0.220 &0.378 &\underline{0.099}&0.088 &\underline{0.022} &0.040 \\
DST-Calib (SB)  &2*$[\pm10^\circ , \pm 1.0\text{m}]$  &0.566&0.134&0.314 &0.345 &0.156&0.086 &0.035 &0.072 
&0.878&0.188&0.360 &0.677 &0.171&0.144 &0.040 &0.080 \\
\textbf{DST-Calib (SB*)}  &2*$[\pm5^\circ , \pm 0.5\text{m}]$  &\textbf{0.257}	&\textbf{0.063}	&\underline{0.082} &\textbf{0.200} &\textbf{0.086}	&\textbf{0.036} &\underline{0.026} &\underline{0.034}
&\textbf{0.295}	&\textbf{0.064}	&0.134  &\underline{0.215} &\textbf{0.055}&\textbf{0.049} &\textbf{0.010}& 0.035 \\

DST-Calib (SB*)  &2*$[\pm10^\circ , \pm 1.0\text{m}]$  &0.429	&0.115	&0.272 &0.261 &0.120&0.077 &0.036 &0.061
&0.745	&0.175	&0.282 &0.593 &0.124&0.134 &0.023 &0.085 \\

DST-Calib (PE)  &2*$[\pm30^\circ , \pm 0.5\text{m}]$  &1.392	&0.176	&0.437 &0.705 &1.015&0.144 &0.057 &0.061
&1.257	&0.148	&0.246 &0.598 &1.016&0.128 &0.058 &\textbf{0.028}\\
DST-Calib (PE+SB*)  &2*$[\pm30^\circ , \pm 0.5\text{m}]$  &0.845	&0.129	&0.110 &0.629 &0.487&0.106 &0.042 &0.036
&0.804	&0.116	&0.225 &0.494 &0.458&0.083 &0.044 &0.048\\

\bottomrule
\end{tabular}
\label{tab.main_kitti00}
\end{table*}

\begin{table*}[t!]
\caption{Comparisons with SoTA LCEC approaches on KITTI Odometry (01-09 sequences). }
\centering
\fontsize{6.6}{10}\selectfont
\begin{tabular}{l|c|c@{\hspace{0.15cm}}c|c@{\hspace{0.15cm}}c|c@{\hspace{0.15cm}}c|c@{\hspace{0.15cm}}c|c@{\hspace{0.15cm}}c|c@{\hspace{0.15cm}}c|c@{\hspace{0.15cm}}c|c@{\hspace{0.15cm}}c}
\toprule
\multirow{2}*{Approach}& \multirow{2}*{Initial Range}&\multicolumn{2}{c|}{01} & \multicolumn{2}{c|}{02} & \multicolumn{2}{c|}{03}  & \multicolumn{2}{c|}{04} & \multicolumn{2}{c|}{ 05} &  \multicolumn{2}{c|}{06}&  \multicolumn{2}{c|}{07} & \multicolumn{2}{c}{09}\\
\cline{3-18}
&& $e_r$ & $e_t$& $e_r$ & $e_t$ & $e_r$ & $e_t$  & $e_r$ & $e_t$ & $e_r$ & $e_t$ & $e_r$ & $e_t$ & $e_r$  & $e_t$ & $e_r$  & $e_t$ \\
\hline
\hline
CRLF \cite{ma2021crlf}  & -
&{0.623}	&7.363
&0.632	&3.642
&0.845	&6.007
&{0.601}&0.372
&0.616	&5.961
&0.615	&25.76
&0.606	&1.807

&0.626	&5.133\\
UMich \cite{pandey2015automatic} & -
&2.196	&{0.305} &3.733
&0.331	&3.201&0.316 &2.086&0.348	&3.526
&0.356 &2.914&0.353	&3.928&0.368 
&3.117	&0.363\\
HKU-Mars \cite{yuan2021pixel}  & -
&20.73	&3.770
&32.95	&12.70
&21.99	&3.493
&4.943	&0.965
&34.42	&6.505
&25.20	&7.437
&33.10	&7.339

&20.38	&3.459\\
DVL \cite{koide2023general} & -
&112.0&2.514
&120.6&4.285	
&124.7&4.711
&113.5&4.871
&123.9&4.286
&128.9&5.408	
&124.7&5.279 
&116.7	&3.931\\

MIAS-LCEC \cite{zhiwei2024lcec}  & -
&0.621	&0.300
&0.801	&0.327
&1.140	&0.324
&0.816	&0.369
&4.768	&0.775
&2.685	&0.534
&11.80	&1.344
&0.998	&0.432
\\
EdO-LCEC \cite{huang2025environment} & -
&2.269 &0.459
 &0.561&0.142
&0.737&0.137 
&1.104&0.339
&\textbf{0.280}&0.093 
&{0.485}&0.124 
&\textbf{0.188}&0.076 
&\underline{0.386}	&0.120 \\

\hline
DST-Calib (DB) & 2*$[\pm5^\circ , \pm 0.5\text{m}]$
&0.706	&0.132	
&0.551	&0.108	
&0.791	&0.128	
&0.576	&0.114	
&0.520	&0.103	
&0.528	&0.110	
&0.528	&0.103	
&0.559	&0.109
\\
DST-Calib (DB) & 2*$[\pm10^\circ , \pm 1.0\text{m}]$
&1.343	&0.262	
&1.198	&0.201	
&1.358	&0.237	
&1.263	&0.208	
&1.108	&0.191	
&1.121	&0.210	
&1.152	&0.191	
&1.207	&0.202
\\
DST-Calib (SB)  & 2*$[\pm5^\circ,\pm 0.5\text{m}]$
&0.651	&0.136
&\underline{0.422}	&\underline{0.088}
&0.617	&\underline{0.113}
&\underline{0.424}	&\underline{0.092}
&0.366	&\underline{0.083}
&\underline{0.397}	&\underline{0.086}
&0.360	&0.077
&0.421	&\underline{0.088}
\\
DST-Calib (SB)  & 2*$[\pm10^\circ, \pm 1.0\text{m}]$
&1.126	&0.263
&0.652	&0.159
&0.849	&0.194
&0.671	&0.183
&0.610	&0.146
&0.611	&0.152
&0.558	&0.135
&0.657	&0.160
\\
\textbf{DST-Calib (SB*)}  & 2*$[\pm5^\circ,\pm 0.5\text{m}]$
&\underline{0.517}	&\textbf{0.084}
&\textbf{0.297}	&\textbf{0.051}
&\textbf{0.439}	&\textbf{0.075}
&\textbf{0.382}	&\textbf{0.068}
&\underline{0.318}	&\textbf{0.058}
&\textbf{0.305}	&\textbf{0.056}
&\underline{0.285}	&\textbf{0.063}
&\textbf{0.273}	&\textbf{0.055}
\\
DST-Calib (SB*)  & 2*$[\pm10^\circ,\pm 1.0\text{m}]$
&0.870	&0.142
&0.440	&0.115
&\underline{0.501}	&0.137
&0.552	&0.162
&0.510	&0.134
&0.525	&0.123
&0.476	&0.097
&0.548	&0.143 
\\
DST-Calib (PE)   & 2*$[\pm30^\circ, \pm 0.5\text{m}]$
&1.365	&0.200
&1.480	&0.178
&2.508	&0.185
&2.313	&0.182
&1.504	&0.222
&0.939	&0.161
&1.497	&0.162
&1.239	&0.195
\\

DST-Calib (PE+SB*)  & 2*$[\pm30^\circ, \pm 0.5\text{m}]$
&\textbf{0.435}	&\underline{0.100}
&0.797	&0.133
&1.700	&0.171
&0.912	&0.105
&0.905	&0.147
&0.637	&0.120
&0.684	&0.109
&0.447	&0.092
\\

\bottomrule
\end{tabular}
\label{tab.cmp_kitti_01_08}
\end{table*}

\subsection{How Training Strategies Affect Model Performance}
\label{sec.exp_how_train_affect}

\subsubsection{The Impact of Data Augmentation Range on Generalization Capability}
\label{sec.exp_data_aug_range}
To further explore why the previous training strategy led to this poor performance and how the data generation range affects generalization, we set different mis-calibration ranges for our proposed double-sided data augmentation and validated the results on the left and right cameras of the KITTI Odometry. To emphasize the impact of the mis-calibration range on data augmentation, we use depth projections of LiDAR point clouds from camera-perspective views to replace the CDP images. The LiDAR projections captured from different views are sent to the DST-Calib DB to obtain their relative poses. We set the $\boldsymbol{c}_{lidar}$ range as $[5^\circ,0.5 \text{m}]$, and the $\boldsymbol{c}_{cam}$ range from $[0^\circ,0\text{m}]$ to $[5^\circ,0.5 \text{m}]$. The results shown in Table \ref{tab.impact_of_different_gen_range} indicate that the generalization capability will increase as the mis-calibration range at the two augmentations' centers becomes larger. When the mis-calibration range of one side is too small, the network will lose its generalization capability and will not be applicable in real-world calibration tasks. While the range $\boldsymbol{c}_{lidar}$ and $\boldsymbol{c}_{cam}$ are similar, the accuracy and generalization ability are promising. This demonstrates that a balanced double-sided data augmentation is necessary for an LCEC network to acquire generalization ability.

\subsubsection{The Impact of Modal Difference of the Model Input}
We also investigate the influence of the modal difference in the input data. In this experiment, we compare the model performance trained by different input setups. As shown in Table \ref{tab.impact_of_diff_input_modal}, the calibration error of the model trained with LDP-LDP input is significantly lower than that of the model trained with LDP-CDP input. This indicates that reducing the modality gap between the two sensor inputs improves calibration performance. Because we employ monocular depth estimation to generate camera depth clouds, the modality gap between LiDAR and camera data is substantially reduced compared with prior approaches, allowing the model to learn more effective cross-modal correlations. Additionally, as illustrated in Fig. \ref{fig.fea_map_cmp} (e), the feature maps extracted by both the DB and SB architectures are noticeably improved compared with those obtained under a single-sided data augmentation strategy. This further validates the effectiveness of the proposed double-sided augmentation in enhancing generalization.

\subsection{Comparison with State-of-the-Art Methods}
\label{sec.exp_cmp_SoTA}

In this section, we compare our proposed DST-Calib with SoTA target-free LCEC methods, including both DNN-based approaches and non-learning-based methods. As shown in Table \ref{tab.main_kitti00}\footnote{The reproduced results of LCCNet yield higher calibration errors compared to those reported in their paper.} and Table \ref{tab.cmp_kitti_01_08}, DST-Calib achieves leading accuracy and strong generalization. Notably, while prior DNN-based methods perform well on the left camera but degrade severely on the right camera, DST-Calib produces highly consistent results across both cameras. 
It significantly outperforms all other DNN-based approaches on the right camera. 
A closer examination reveals that when the initial mis-calibration range is relatively small (e.g., $[\pm 5^\circ, 0.5\text{m}]$), the translation error of previous DNN-based methods on the right camera is about half of the imposed perturbation (approximately 0.3-0.4m). As the initial mis-calibration increases, the error further increases to roughly 0.4-0.6m (almost equal to the physical distance between the left and right cameras). This clearly indicates that these methods overfit to the specific extrinsic parameters seen during training. The seemingly high accuracy of these DNN-based methods on the KITTI left camera is misleading, as the trained models fail to generalize even to another camera position within the same scene of the same dataset.

According to experimental results across all KITTI Odometry test sequences (00–09), the proposed DST-Calib achieves substantially more stable performance than other SoTA target-free LCEC methods. Moreover, under the self-supervised pathway, DST-Calib PE attains promising rotation calibration accuracy even when initialized with a large rotational deviation of up to $30^\circ$ and a translational deviation of up to 1 m. When both pathways are activated, the PE+SB* configuration yields consistently accurate results for both rotation and translation across very large mis-calibration ranges, highlighting the effectiveness of the proposed dual-path calibration framework built on double-sided data augmentation.
Furthermore, the overall performance of the SB architecture is better than that of the DB architecture. SB-based models converge more easily and generalize better to unseen scenarios. This confirms that the proposed single-branch design enhances feature correlation and pose regression by leveraging the explicitly constructed difference map, which strengthens feature reliability and cross-modal association. Additionally, multi-frame optimization provides a significant accuracy boost. For example, on sequence 01, applying multi-frame optimization reduces the rotation error of SB by approximately 20.6\%-22.7\% and the translation error by approximately 38.2\%-46.0\%. We attribute this improvement to the scoring mechanism used in multi-frame optimization, which effectively enhances the final extrinsic calibration accuracy by averaging rotation and translation estimates across frames.

\subsection{Performance of Depth Anchor Correction}
\begin{table}[t!]
\caption{
Quantitative comparisons of SoTA Monocular Depth Estimation on KITTI Dataset. }
\centering
\fontsize{6.7}{10}\selectfont
\begin{tabular}{l|c@{\hspace{0.15cm}}c@{\hspace{0.15cm}}c@{\hspace{0.15cm}}c|c@{\hspace{0.15cm}}c@{\hspace{0.15cm}}c}
\toprule
\multirow{2}*{Approach} & \multicolumn{4}{c|}{The lower the better} & \multicolumn{3}{c}{The higher the better}\\
\cline{2-8}
& Abs Rel &Sq Rel &RMSE &RMSE log &$\delta_1$ &$\delta_2$ &$\delta_3$\\
\hline
\hline
LEGO \cite{yang2018lego} & 0.192 &1.352 &6.276 &0.252 &0.783 &0.921 &0.969  \\
MonoDepth2 \cite{godard2019digging}&0.106 &0.818 &4.750 &0.196& 0.874 &0.957 &0.979 \\
Fusion Depth \cite{feng2022advancing} & \textbf{0.063} &0.364 &3.291 &0.139 &0.945 &0.978 &0.988\\
DPA2 \cite{depth_anything_v2} &0.521&3.220&4.558&0.546&0.313&0.578&0.789\\
MoGe2 \cite{wang2025moge}& 0.206 &0.480 &2.240 &0.265 &0.517	&0.963	&0.985\\
\hline
DPA2+DAR &\underline{0.087} &\textbf{0.149} &\textbf{1.191}	&\textbf{0.119} &\textbf{0.950} &\textbf{0.987} &\textbf{0.994}\\
MoGe2+DAR &0.089&\underline{0.169} &\underline{1.281} &\underline{0.121} &\underline{0.947}&\underline{0.986}&\textbf{0.994}\\

\bottomrule
\end{tabular}
\label{tab.dca_sota_cmp}
\end{table}
\begin{table}[t!]
\caption{
Quantitative comparisons of Depth Correction.}
\centering
\fontsize{6.3}{10}\selectfont
\begin{tabular}{c|l|c@{\hspace{0.15cm}}c@{\hspace{0.15cm}}c@{\hspace{0.15cm}}c|c@{\hspace{0.15cm}}c@{\hspace{0.15cm}}c}
\toprule

Dataset& Approach& MAE &RMSE &Abs Rel &Sq Rel&$\delta_1$ &$\delta_2$ &$\delta_3$\\
\hline
\hline
\multirow{4}*{KITTI-360}&
DPA2 &5.352	&5.931 &0.830	&6.059	&0.12	&0.364	&0.666\\
&DPA2+DAR &0.786 &\underline{1.353} &\underline{0.100}	&\underline{0.247}	&\underline{0.908}	&\underline{0.976}	&\underline{0.991}\\
&MoGe2 &\underline{1.378}&1.783&0.174&0.353&0.711&0.962&0.978 \\
&MoGe2+DAR &\textbf{0.675}&\textbf{1.126}&\textbf{0.085}&\textbf{0.159}&\textbf{0.931}&\textbf{0.979}&\textbf{0.992}\\
\hline
\multirow{4}*{Argoverse2}&
DPA2 &12.69	&12.71 &2.569	&32.60	&0.000	&0.001	&0.008\\
&DPA2+DAR &\underline{0.950} &\underline{1.546} &\textbf{0.078}	&\underline{0.232}	&0.928	&\underline{0.981}	&\textbf{0.995}\\
&MoGe2 &1.265	&1.577 &0.098	&\textbf{0.192}	&\textbf{0.961}	&\textbf{0.986}	&\underline{0.994} \\
&MoGe2+DAR &\textbf{0.923}	&\textbf{1.405} &\underline{0.088}	&0.270	&\underline{0.942}	&0.979	&0.991\\
\hline
\multirow{4}*{nuScenes}&
DPA2 &3.669	&3.950	&0.369	&1.551	&0.232	&0.887	&0.989\\
&DPA2+DAR &\underline{0.766}	&\underline{1.136}	&\underline{0.066}	&\underline{0.110}	&\underline{0.980}	&\textbf{0.998}	&\textbf{1.000}\\
&MoGe2 &1.725	&2.070	&0.144	&0.310	&0.861	&0.994	&0.997\\
&MoGe2+DAR &\textbf{0.591}	&\textbf{0.908}	&\textbf{0.050}	&\textbf{0.066}	&\textbf{0.992}	&\underline{0.996}	&\textbf{1.000}\\
\hline
\multirow{4}*{TF70}&
DPA2 &6.155	&6.856 &1.502	&16.96	&0.178	&0.393	&0.578\\
&DPA2+DAR &\textbf{0.795}	&\textbf{1.478} &\textbf{0.102}	&\textbf{0.486}	&\textbf{0.916}	&\textbf{0.963}	&\textbf{0.984}\\
&MoGe2 &1.340	&1.781 &0.184	&\underline{0.574}	&0.749	&0.939	&\underline{0.981}\\
&MoGe2+DAR &\underline{0.983}	&\underline{1.565} &\underline{0.134}	&0.590	&\underline{0.898}	&\underline{0.958}	&0.980\\

\bottomrule
\end{tabular}
\label{tab.dca_robust_test}
\end{table}

In this section, we evaluate the performance of the proposed depth refinement method, DAR. We first compare the depth-correction accuracy against representative monocular depth estimation approaches on the KITTI dataset. As shown in Table \ref{tab.dca_sota_cmp}, our method achieves superior accuracy, markedly improving over the initial depth maps produced by Depth Anything V2 and MoGe2. Moreover, the refined depth results significantly surpass those of SoTA self-supervised monocular metric depth estimation methods, including LEGO \cite{yang2018lego}, MonoDepth2 \cite{godard2019digging}, and FusionDepth \cite{feng2022advancing}. These substantial performance gains clearly demonstrate the effectiveness and practical value of the proposed depth correction strategy. To further assess robustness, we additionally evaluate DAR on KITTI-360, Argoverse2, and TF70. The quantitative results in Table \ref{tab.dca_robust_test} show that our method consistently improves the raw depth predictions obtained from large vision models, producing camera depth maps that more closely resemble true LiDAR measurements. In particular, MoGe2+DAR attains an average MAE of 0.793 and an average RMSE of 1.251 across the four datasets, indicating stable depth correction across diverse environments and sensor setups.

\begin{figure*}[t!]
    \centering
    \includegraphics[width=0.99\linewidth]{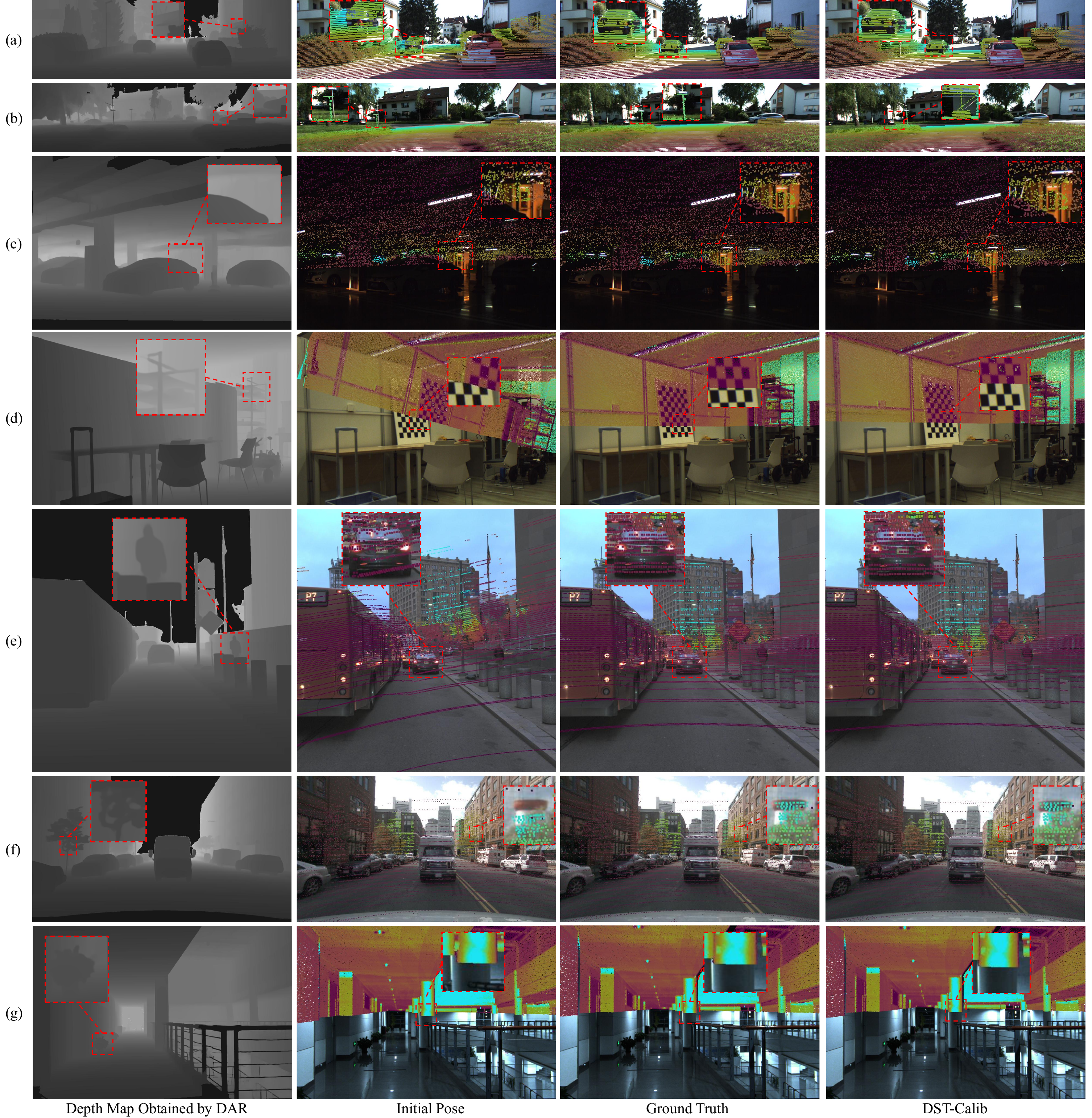}
    \caption{Qualitative results of DST-Calib on diverse real-world datasets: (a)-(g) the visualization of LiDAR-Camera data fusion on KITTI Odometry, KITTI-360, MIAS-LCEC-TF360, MIAS-LCEC-TF70, Argoverse2, nuScenes, and LCScenes. For visualization, all LiDAR projections are overlaid with their respective RGB images using the calibrated extrinsic parameters. Monocular depth maps obtained by DAR are also provided.}
    \label{fig.cross_domain_visual}
\end{figure*}
\begin{table*}[t!]
\caption{Quantitative comparisons with SoTA non-learning-based target-free LCEC approaches on MIAS-LCEC Datasets. The results of DST-Calib are zero-shot generalization results.}
\centering
\fontsize{6.5}{10}\selectfont
\begin{tabular}{l|c|c@{\hspace{0.15cm}}c|c@{\hspace{0.15cm}}c@{\hspace{0.15cm}}c|c@{\hspace{0.15cm}}c@{\hspace{0.15cm}}c|c@{\hspace{0.15cm}}c|c@{\hspace{0.15cm}}c@{\hspace{0.15cm}}c|c@{\hspace{0.15cm}}c@{\hspace{0.15cm}}c}
\toprule
\multirow{3}*{Approach}& \multirow{3}*{Initial Range}&\multicolumn{8}{c|}{MIAS-LCEC-TF70} &\multicolumn{8}{c}{MIAS-LCEC-TF360}\\
\cline{3-18}
&&\multicolumn{2}{c|}{Magnitude}
&\multicolumn{3}{c|}{Rotation Error ($^\circ$)} &\multicolumn{3}{c|}{Translation Error (m)} 
&\multicolumn{2}{c|}{Magnitude}
&\multicolumn{3}{c|}{Rotation Error ($^\circ$)} &\multicolumn{3}{c}{Translation Error (m)}\\

&& $e_r$ ($^\circ$) & $e_t$ (m) & Yaw & Pitch & Roll  & {X} &  {Y} &  {Z}   & $e_r$ ($^\circ$) & {$e_t$ (m)} & Yaw & Pitch & Roll   &  {X} &  {Y} &  {Z}\\
\hline
\hline

CRLF \cite{ma2021crlf} &- &1.683	&11.13	&0.197	&1.625	&0.946	&10.05	&3.036	&2.589 
&1.463	&7.418	&0.135	&1.421	&0.317	&7.093	&0.621	&1.885\\
UMich \cite{pandey2015automatic}&- &4.265	&0.333 &0.485	&1.945	&4.272	&0.217	&0.134	&0.115	
&4.434	&0.249	&\underline{0.324}	&2.340	&3.356	&0.137	&0.129	&0.110	\\
HKU-Mars \cite{yuan2021pixel} &- &3.941	&1.261	&2.140	&2.156	&1.988	&0.806	&0.555	&0.475 
&56.41&	8.042	&33.45	&35.59	&7.843	&3.646	&5.931	&2.227\\
DVL \cite{koide2023general}  &- &0.423	&0.100	&0.201	&0.292	&0.104	&\underline{0.075}	&0.050	&0.026 
&54.70	&1.325	&23.98	&43.27	&34.73&	0.335	&0.769	&0.892\\
MIAS-LCEC \cite{zhiwei2024lcec}  &- &\underline{0.298}	&\underline{0.061}	&\underline{0.133}	&\underline{0.196}	&\underline{0.110}	&\textbf{0.040}	&\underline{0.028}	&\underline{0.019}
&0.799	&0.142	&0.349	&0.535	&0.277	&0.095	&0.075	&0.050\\
EdO-LCEC \cite{huang2025environment}  &- &\textbf{0.255}	&\textbf{0.055}	&\textbf{0.117}	&\textbf{0.166}	&\textbf{0.096}	&\textbf{0.040}	&\textbf{0.024}	&\textbf{0.016}
&\textbf{0.504}	&\textbf{0.107}	&\textbf{0.229}	&\underline{0.341}	&\textbf{0.112}	&\textbf{0.064}	&0.066	&\textbf{0.032}\\

\hline
DST-Calib (SB)  &2*$[\pm5^\circ,\pm 0.5m]$ &1.200	&0.134	&0.920 &0.424 &0.337&0.095 &0.049  &0.049
&1.232	&0.134	&0.953 &0.364 &0.392 &0.094  &\underline{0.056} &\underline{0.047} \\
DST-Calib (SB*)  &2*$[\pm5^\circ,\pm 0.5m]$ &0.841	&0.143	&0.518 &0.382 &0.371&0.113 &0.054 &0.044
&\underline{0.667}	&\underline{0.129}	&0.419 &\textbf{0.288} &\underline{0.274}&\underline{0.079} &0.062 &0.058\\
DST-Calib (PE)  &2*$[\pm30^\circ,\pm 0.5m]$  &1.903&0.181&0.513 &0.882 &1.453&0.145 &0.054 &0.053
&2.303&0.152&2.068 &0.510 &0.684&0.117 &\textbf{0.049} &0.053\\
DST-Calib (PE+SB*)  &2*$[\pm30^\circ,\pm 0.5m]$  &1.423&0.149&0.549 &0.654 &1.021&0.100 &0.060 &0.071
&1.871&0.165&1.656 &0.662 &0.300&0.122 &0.080 &0.053\\
\bottomrule
\end{tabular}
\label{tab.mias_lcec}
\end{table*}

\begin{figure*}[t!]
    \centering
    \includegraphics[width=1\linewidth]{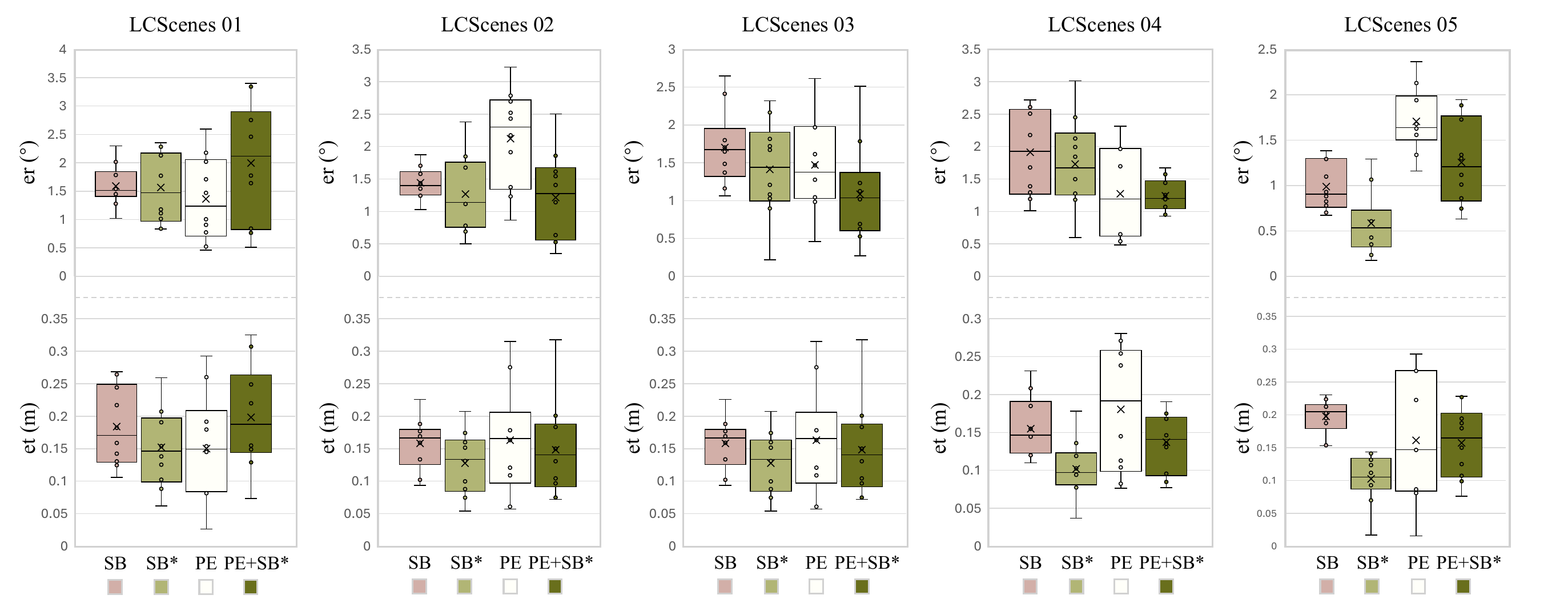}
    \caption{Quantitative calibration results on our created LCScenes dataset. The initial mis-calibration range of all architectures is set to 2*$[5^\circ,0.5\text{m}]$.}
    \label{fig.eva_LCScenes}
\end{figure*}

\begin{table*}[t!]
\caption{Zero-shot generalization results on three public datasets with various sensor configurations.}
\centering
\fontsize{6.7}{10}\selectfont
\begin{tabular}{l|c|cc|cc|cc|cc|cc|cc}
\toprule
\multirow{2}*{Approach}& \multirow{2}*{Initial Range} & \multicolumn{2}{c|}{\makecell{KITTI-360 \\ SFL + SFR}} & \multicolumn{2}{c|}{\makecell{Argoverse2 \\ SFL + SFR + RFR}}& \multicolumn{2}{c|}{\makecell{Argoverse2 \\ RSL + RSR }}  & \multicolumn{2}{c|}{\makecell{Argoverse2 \\ RBL + RBR}}  &  \multicolumn{2}{c|}{\makecell{nuScenes \\ RFC + RFL + RFR }} &  \multicolumn{2}{c}{\makecell{nuScenes \\ RBC + RBL + RBR}}\\
\cline{3-14}
& & $e_r$ ($^\circ$) & $e_t$ (m)& $e_r$ ($^\circ$) & $e_t$ (m)& $e_r$ ($^\circ$) & $e_t$ (m)& $e_r$ ($^\circ$) & $e_t$ (m)& $e_r$ ($^\circ$) & $e_t$ (m)& $e_r$ ($^\circ$) & $e_t$ (m)\\
\hline
\hline
DST-Calib (SB)  &2*$[\pm5^\circ, \pm 0.5\text{m}]$
&\underline{0.898}	&0.137
&\underline{0.938}	&\underline{0.142}
&1.336	&\underline{0.154}
&0.901	&\textbf{0.141}
&1.332	&0.174
&1.333	&0.165
\\
DST-Calib (SB*)  &2*$[\pm5^\circ, \pm 0.5\text{m}]$    
&\textbf{0.709}	&\textbf{0.113}
&\textbf{0.803}	&\textbf{0.129}
&0.892	&\textbf{0.151}
&\underline{0.733}	&\underline{0.148}
&\textbf{0.880}	&\textbf{0.135}
&\textbf{0.954}	&\textbf{0.096}
 \\

DST-Calib (SB)  &2*$[\pm15^\circ, \pm 0.5\text{m}]$
&3.980	&0.169
&3.940	&0.187
&4.503	&0.180
&3.965	&0.187
&4.281	&0.188
&4.340	&0.187 \\
\hline
DST-Calib (PE)  &2*$[\pm5^\circ,\pm 0.5\text{m}]$
&2.173	&0.208
&1.579	&0.164
&\textbf{0.807}	&0.208
&1.206	&0.179
&\underline{1.144}	&0.192
&1.161	&0.181 \\
DST-Calib (PE)  &2*$[\pm15^\circ,\pm 0.5\text{m}]$
&2.014	&0.169
&1.070	&0.163
&1.099	&0.164
&1.109	&0.205
&1.399	&0.172
&1.599	&0.184 \\
DST-Calib (PE)  &2*$[\pm30^\circ,\pm 0.5\text{m}]$
&2.105	&0.153
&1.441	&0.171
&1.421	&0.202
&1.083	&0.174
&1.383	&0.160
&1.266	&0.178 \\

DST-Calib (PE+SB*) &2*$[\pm5^\circ,\pm 0.5\text{m}]$ 
&1.804	&0.154
&1.049	&0.189
&0.956	&0.180
&{0.885}	&0.161
&1.289	&0.148
&\underline{0.992}	&\underline{0.138} \\

DST-Calib (PE+SB*) &2*$[\pm30^\circ,\pm 0.5\text{m}]$ 
&1.824	&\underline{0.136}
&1.230	&0.179
&\underline{0.831}	&0.166
&\textbf{0.688}	&0.151
&1.155	&\underline{0.142}
&1.112	&0.161 \\
\bottomrule
\end{tabular}
\label{tab.cross_domain_generalization}
\end{table*}
\begin{figure*}[t!]
    \centering
    \includegraphics[width=0.99\linewidth]{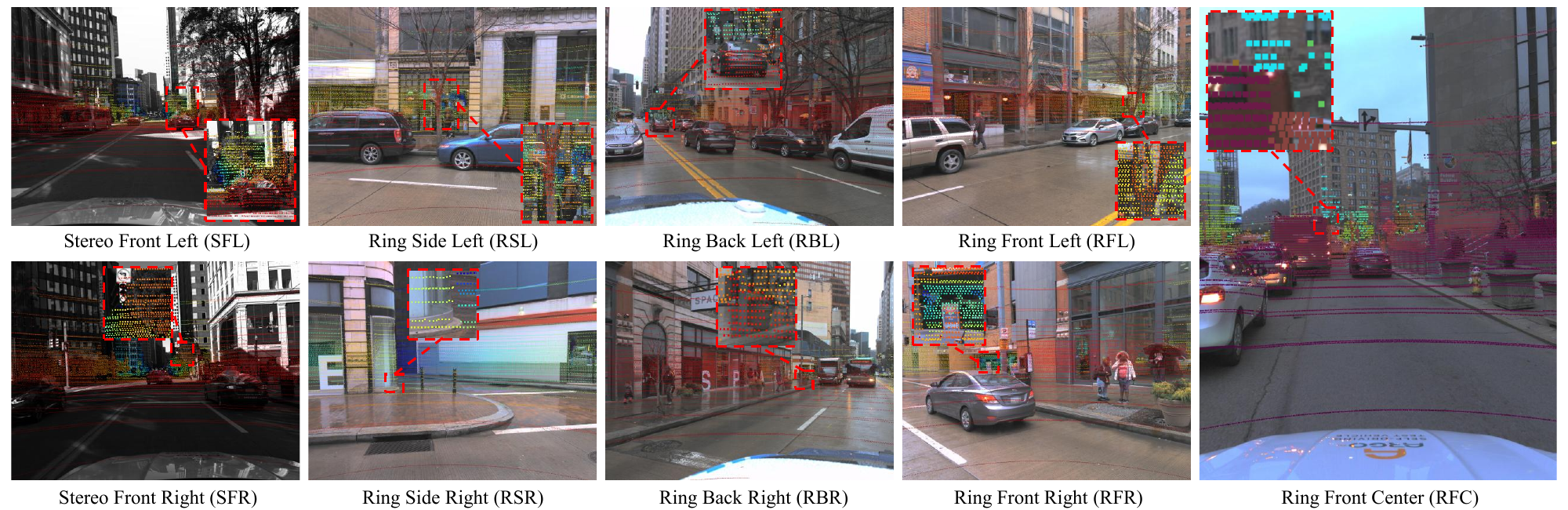}
    \caption{Visualization of data fusion results on all cameras of the Argoverse2 dataset. LiDAR point clouds are projected onto their respective RGB images using the extrinsic parameters calibrated by DST-Calib.}
    \label{fig.calibvisual_Argo2MultiCamera}
\end{figure*}

\subsection{Cross-Domain Generalization}

To further assess generalization, we test the proposed DST-Calib on the other four additional public datasets, as well as our LCScenes dataset, covering a total of 19 distinct LiDAR-camera configurations. Since the DB architecture is not our main contribution and its performance is generally inferior to that of SB (as shown in Table \ref{tab.main_kitti00} and Table \ref{tab.cmp_kitti_01_08}), we do not include DB in the evaluation of cross-domain generalization. Notably, DST-Calib is trained only on the front-view cameras from KITTI and Argoverse; all other cameras, LiDARs, and scene domains remain completely unseen during training. Consequently, the reported calibration results constitute a zero-shot evaluation, posing a highly challenging scenario for any target-free LCEC method to achieve accurate extrinsic estimation. 

First, we evaluate the cross-domain generalization capability of DST-Calib on MIAS-LCEC-TF70 and MIAS-LCEC-TF360. These two datasets contain point clouds captured by different types of solid-state LiDARs with distinct scanning fields of view ($70^\circ$ and $360^\circ$). Since solid-state LiDAR data are never used during training, calibrating their extrinsic parameters presents a considerable challenge. Nevertheless, the experimental results in Table \ref{tab.mias_lcec} exceed our expectations. Although the calibration accuracy is slightly inferior to that of MIAS-LCEC and EdO-LCEC, DST-Calib outperforms other target-free LCEC approaches specifically designed for dense point clouds from solid-state LiDARs. Moreover, when adopting the self-supervised pathway, DST-Calib still achieves higher calibration accuracy than CRLF, UMich, and HKU-Mars, even though these methods are evaluated under a much smaller mis-calibration range (within 1° and 0.1 m).

Second, we evaluate DST-Calib on LCScenes. The quantitative results for each sequence are shown in Fig. \ref{fig.eva_LCScenes}. Unlike the MIAS-LCEC dataset, LCScenes includes more indoor environments and contains varying camera resolutions and sensor configurations. 
Experimental results show that all DST-Calib architectures maintain stable performance across the five sequences. Except for sequence 01, SB* consistently outperforms SB, and PE+SB* always achieves higher calibration accuracy than PE, demonstrating the effectiveness of multi-frame optimization in improving both accuracy and robustness. Additionally, although the self-supervised PE is less stable than the fully-supervised SB, its average calibration error remains comparable. This confirms that our self-supervised strategy, enabled by double-sided data augmentation, is both effective and reliable. Under a relatively small initial mis-calibration range, the performance of the fully-supervised DST-Calib pathway is better than that of the self-supervised pathway. With guidance from extrinsic ground truth, the model can learn the uncertain part of point cloud similarity that is not accessible to self-supervision.

Finally, we further validate DST-Calib on Argoverse2, nuScenes, and KITTI-360, whose LiDAR point clouds are collected using mechanical spinning LiDARs. Notably, Argoverse2 and nuScenes include cameras facing the left, right, and rear of the vehicle, which differ significantly from the front-facing cameras used during training. A total of 15 sensor combinations are evaluated to thoroughly examine the zero-shot generalization capability. The quantitative results are summarized in Table \ref{tab.cross_domain_generalization}, and qualitative examples are visualized in Fig. \ref{fig.calibvisual_Argo2MultiCamera}. Across these heterogeneous datasets, DST-Calib maintains low calibration error and demonstrates strong adaptability to diverse environments and challenging illumination conditions. These promising results indicate that the network successfully learns the underlying relationship between the input projections and their corresponding extrinsic parameters. We attribute this improvement to the proposed single-branch architecture with explicit difference-map construction, which replaces the ambiguous feature-correlation process of previous DNN-based approaches with a clear, deterministic association between the two modalities. 

Fig. \ref{fig.cross_domain_visual} summarizes the overall visualization results across all datasets. The fused LiDAR-camera projections obtained using the calibrated extrinsic parameters exhibit clear alignment along both geometric structures and textural edges. Although DST-Calib does not achieve the same level of accuracy in certain indoor environments (where the point cloud depth range is inherently limited), it still delivers significantly improved alignment compared to the initial pose, even when the initial rotation error is large.

\subsection{Computational Cost and Runtime Analysis}
\label{sec.exp_runtime}

\begin{table}[t!]
\caption{
Computational cost and runtime analysis of DST-Calib.}
\centering
\fontsize{6.7}{10}\selectfont
\begin{tabular}{l|c@{\hspace{0.15cm}}c@{\hspace{0.15cm}}c@{\hspace{0.15cm}}c|c}
\toprule
{Approach} & \makecell{Depth \\ Estimation \\(s/frame)} & \makecell{Depth \\Correction\\(s/frame)} &\makecell{Model \\ Inferene \\(s/frame)} & \makecell{Pose \\Optimization \\ (s/frame)} & \makecell{Total\\(s/frame)}\\
\hline
\hline
DST-Calib (DB) & 0.661 &0.052 &0.009 &0.066 &0.871  \\
DST-Calib (SB) &0.598 &0.053 &0.007 &0.030& 0.772 \\
DST-Calib (PE) & 0.640 &0.050 &0.007 &0.027 &0.808 \\
DST-Calib (PE+SB*) &0.634 &0.051&0.013&0.055&0.869 \\

\bottomrule
\end{tabular}
\label{tab.exp_runtime}
\end{table}

Our algorithm is implemented on an Intel i7-14700K CPU and an NVIDIA RTX4090D GPU. We evaluate the computational cost of each major component in the calibration pipeline. The total processing time for a single frame, including depth estimation, depth correction, model inference, and pose optimization, is approximately 0.8 seconds. The single-frame processing time should be multiplied by the number of frames used for calibration to obtain the total runtime of the multi-frame optimization. As summarized in Table \ref{tab.exp_runtime}, the depth correction, model inference, and pose optimization stages are highly efficient, requiring only about 0.09 seconds in total. The primary bottleneck lies in the monocular depth estimation stage. When using large-scale models such as Depth Anything V2 or MoGe2, the per-frame processing time increases to roughly 0.65 seconds. Nevertheless, considering that most LiDAR sensors operate at 10-20 Hz, the overall computational cost remains sufficiently low for online calibration. 

\subsection{Ablation Study and Analysis}
\label{sec.ablation_study}
\begin{table}[t!]
\caption{
Quantitative comparisons of different feature extraction backbones with different block sizes (using our novel single-branch architecture of DST-Calib).}
\centering
\settablefont
\begin{tabular}{c|c|rr|rr}
\toprule
\multirow{2}*{Backbone} & \multirow{2}*{Block Size} & \multicolumn{2}{c|}{KITTI Left00} & \multicolumn{2}{c}{KITTI Right00}\\
& &$e_r$ ($^\circ$) & $e_t$ (m)  & $e_r$ ($^\circ$) & $e_t$ (m)\\
\hline
\hline
\multirow{3}*{ResNet 18}  
& 1 &\textbf{0.282}&0.068&\textbf{0.488}&\underline{0.111}\\
& 3 &0.351&0.075&0.576&0.123\\
& 5 &0.366 &0.078&\underline{0.505}&\textbf{0.109}\\
\hline

\multirow{3}*{ResNet 34}  
& 1 &0.316&0.076&0.544&\textbf{0.109}\\
& 3 &0.387&0.090&0.524&0.116\\
& 5 &0.402&0.089&0.508&0.117\\
\hline

\multirow{3}*{Efficient Net}  
& 1 &0.386&0.079&0.514&0.120\\
& 3 &\underline{0.310}&\underline{0.064}&0.544&0.112\\
& 5 &0.311&\textbf{0.062}&0.591&0.137\\

\bottomrule
\end{tabular}
\label{tab.impact_of_different_backbone_blocksize}
\end{table}
\begin{table}[t!]
\caption{
Performance Comparison between different architectures of the pose estimator. }
\centering
\settablefont
\begin{tabular}{c|c@{\hspace{0.15cm}}c@{\hspace{0.15cm}}c|r@{\hspace{0.15cm}}r|r@{\hspace{0.15cm}}r}
\toprule
\multirow{2}*{Architecture} & \multirow{2}*{$\mathcal{L}_{CD}$} & \multirow{2}*{$\mathcal{L}_{t_{ini}}$}& \multirow{2}*{$\mathcal{L}_{eva}'$}& \multicolumn{2}{c|}{KITTI Left00} & \multicolumn{2}{c}{KITTI Right00}\\
& & & &$e_r$ ($^\circ$) & $e_t$ (m)  & $e_r$ ($^\circ$) & $e_t$ (m)\\
\hline
\hline
\multirow{4}*{PE (Simple)}
&\checkmark & &
&1.181&0.581&1.466&0.669\\ 
&\checkmark & \checkmark &
&0.897 &\underline{0.162} &1.197&0.205\\ 
&\checkmark &  & \checkmark
&1.333&0.600&1.438&0.694\\ 
&\checkmark & \checkmark & \checkmark
&0.919&0.182&1.189&0.203\\ 
\hline
\multirow{4}*{PE (Standard)}
&\checkmark & &
&1.398&0.608&1.672&0.691\\ 
&\checkmark & \checkmark &
&1.187&0.208 &\underline{1.132} &\underline{0.191}\\ 
&\checkmark &  & \checkmark
&\textbf{0.849}&0.388&1.260&0.658\\ 
&\checkmark & \checkmark & \checkmark
&\underline{0.890}&\textbf{0.135}&\textbf{1.086}&\textbf{0.181}\\ 
\bottomrule
\end{tabular}
\label{tab.cmp_pe}
\end{table}

\subsubsection{Influence of Feature Extraction and Block Size}
The DB and SB of DST-Calib both contain a feature extraction module and a block processor for pose regression (which is different from previous double-branch networks). To evaluate the function of these two modules. we test the SB architecture of DST-Calib on KITTI left and right cameras with different feature extraction backbones and test different block sizes. Table \ref{tab.impact_of_different_backbone_blocksize} demonstrates that the performance under different feature extraction backbones and different grid sizes achieves similar calibration accuracy. This further proves that, concerning the DNN-based LCEC network that directly regresses the extrinsic parameters, the detailed network modules are not essential. What really matters is the training strategy and the overall feature correlation architecture. It is the double-sided data augmentation and the single-branch architecture of DST-Calib that greatly improve the generalization ability of the LCEC network, but not a specific feature extraction module or the downstream pose regression layers.

\subsubsection{The Performance Comparison between Different Architectures of the Pose Estimator}

We compare the calibration performance of the simple and standard architectures (introduced in Fig. \ref{fig.ComponentsOfZVNet}) of the pose estimator, and further examine the contribution of each loss component in the self-supervised pathway. As shown in Table \ref{tab.cmp_pe}, both architectures exhibit similar performance on the KITTI left and right cameras under all loss-term combinations, while the standard architecture (using LDP and CDP as inputs) achieves slightly higher overall accuracy. Among all loss terms, the Chamfer distance plays the most critical role, substantially reducing rotation errors. When combined with the initial pose constraint, the pose estimator achieves strong rotation-only calibration while preserving the initial translation of the provided extrinsic guess. Incorporating the $\mathcal{L}_{eva}'$ term further improves accuracy when used together with the other two losses.

Since the pose estimator assumes that extrinsic parameters remain consistent throughout the entire sequence, in real-world applications, a simple architecture is sufficient to achieve good rotation-only calibration with fewer model parameters. The combination of the pose estimator and the evaluation module (DST-Calib PE+SB*) is the best choice for robust, accurate 6-DoF extrinsic parameter estimation.

\subsubsection{The Adaptability to Different Mis-Calibration Range on Rotation}
\label{sec.exp_ablation_adapt_miscalibration_range}
\begin{figure}[t!]
    \centering
    \includegraphics[width=0.987\linewidth]{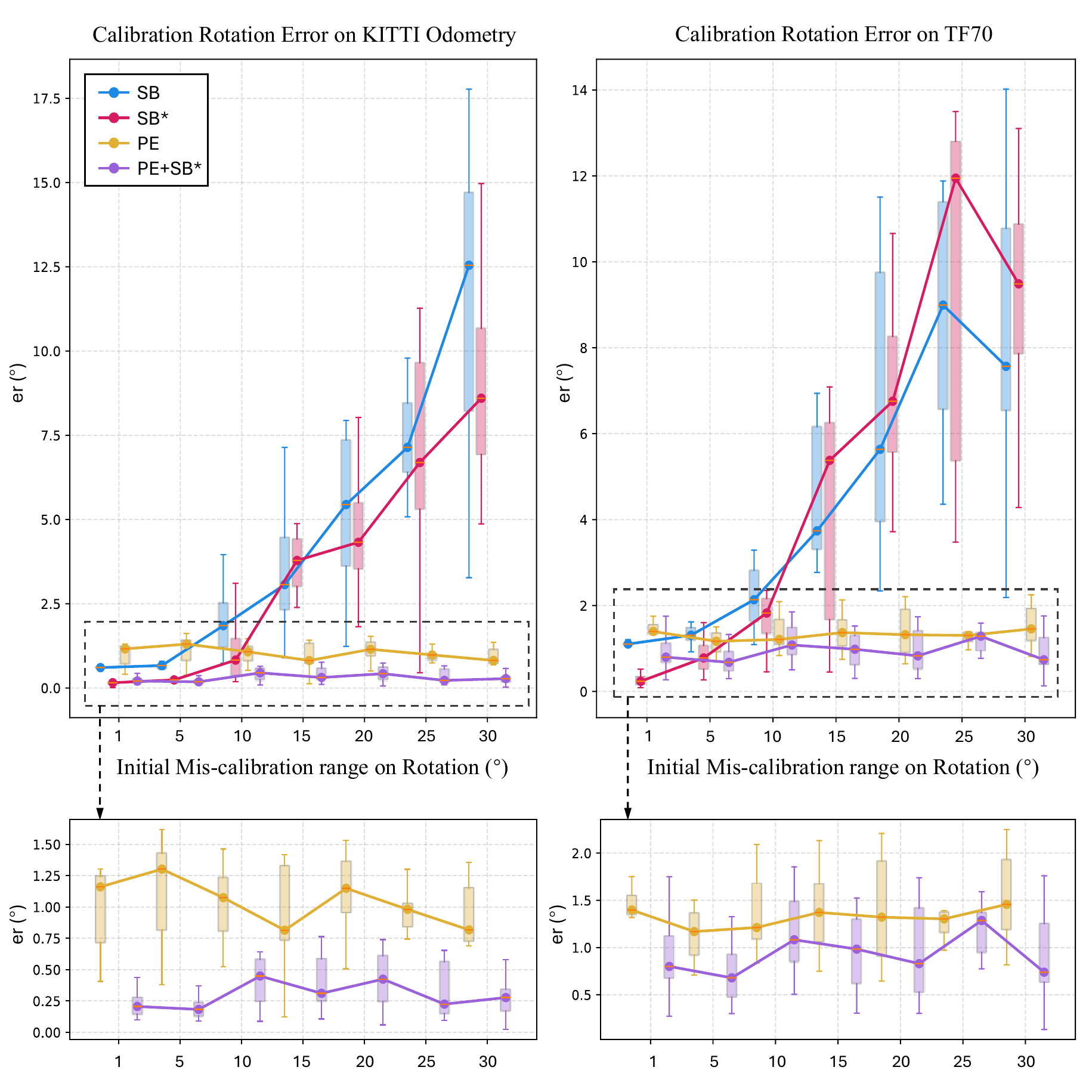}
    \caption{Calibration error on rotation under different initial mis-calibration ranges from 1 to 30 degrees.}
    \label{fig.Rrange}
\end{figure}
In this ablation study, we evaluate the adaptability of DST-Calib to different rotation mis-calibration ranges. We initialize the extrinsic parameters with rotation perturbations varying from a small deviation of $1^\circ$ up to a large $30^\circ$, and measure the resulting calibration accuracy to assess the robustness of each architecture. As shown in Fig. \ref{fig.Rrange}, the results are consistent with expectations. For the SB architecture, the average rotation error increases once the initial deviation exceeds the range seen during training; although multi-frame optimization helps reduce errors within the trained range, it struggles when the initial mis-calibration becomes too large. In contrast, when the self-supervised pathway is activated, both PE and PE+SB* maintain stable and consistent performance across all mis-calibration levels. This demonstrates that the online self-supervised learning significantly improves the robustness to large rotation deviations. Furthermore, PE+SB* consistently outperforms PE alone, indicating that combining the two pathways provides the highest accuracy and the most reliable performance for rotation-only calibration under varying mis-calibration ranges.

\section{Conclusion}
\label{sec.conclusion}
In this article, we dive deeper into the DNN-based target-free LiDAR-camera extrinsic calibration network. We identify a degradation in generalization ability in previous approaches that arises from the single-sided data augmentation, and address it with a novel double-sided data augmentation strategy. Unlike previous approaches, our method is no longer tied to specific sensor configurations and can generalize across diverse environments and a wide range of sensor pairs. To further enhance robustness and accuracy, we introduce a dual-path calibration framework that integrates both fully-supervised and self-supervised pathways, enabling completely target-free calibration and greatly reducing the reliance on pre-calibrated extrinsic ground truth during training. Extensive evaluations on multiple public real-world datasets, as well as our newly collected dataset, demonstrate that our method achieves state-of-the-art performance.

\normalem

\begin{thebibliography}{10}
\providecommand{\url}[1]{#1}
\csname url@samestyle\endcsname
\providecommand{\newblock}{\relax}
\providecommand{\bibinfo}[2]{#2}
\providecommand{\BIBentrySTDinterwordspacing}{\spaceskip=0pt\relax}
\providecommand{\BIBentryALTinterwordstretchfactor}{4}
\providecommand{\BIBentryALTinterwordspacing}{\spaceskip=\fontdimen2\font plus
\BIBentryALTinterwordstretchfactor\fontdimen3\font minus \fontdimen4\font\relax}
\providecommand{\BIBforeignlanguage}[2]{{%
\expandafter\ifx\csname l@#1\endcsname\relax
\typeout{** WARNING: IEEEtran.bst: No hyphenation pattern has been}%
\typeout{** loaded for the language `#1'. Using the pattern for}%
\typeout{** the default language instead.}%
\else
\language=\csname l@#1\endcsname
\fi
#2}}
\providecommand{\BIBdecl}{\relax}
\BIBdecl

\bibitem{giulia2025intergrat}
G.~Pagnanelli \emph{et~al.}, ``Integrating human-like impedance regulation and model-based approaches for compliance discrimination via biomimetic optical tactile sensors,'' \emph{IEEE Transactions on Robotics}, vol.~41, pp. 857--870, 2025.

\bibitem{cattaneo2025cmrnext}
D.~Cattaneo and A.~Valada, ``{CMRNext}: Camera to lidar matching in the wild for localization and extrinsic calibration,'' \emph{IEEE Transactions on Robotics}, vol.~41, pp. 1995--2013, 2025.

\bibitem{zheng2025fastlivo2}
C.~Zheng \emph{et~al.}, ``{FAST-LIVO2}: Fast, direct {LiDAR}–inertial–visual odometry,'' \emph{IEEE Transactions on Robotics}, vol.~41, pp. 326--346, 2025.

\bibitem{li2022deepfusion}
Y.~Li \emph{et~al.}, ``{DeepFusion}: {LiDAR-camera} deep fusion for multi-modal {3D} object detection,'' in \emph{Proceedings of the IEEE/CVF Conference on {Computer Vision and Pattern Recognition} (CVPR)}, 2022, pp. 17\,182--17\,191.

\bibitem{tian2022kimera}
Y.~Tian \emph{et~al.}, ``Kimera-multi: Robust, distributed, dense metric-semantic slam for multi-robot systems,'' \emph{IEEE Transactions on Robotics}, vol.~38, no.~4, 2022.

\bibitem{zhang2024toward}
A.~Zhang \emph{et~al.}, ``Toward robust robot {3-D} perception in urban environments: The ut campus object dataset,'' \emph{IEEE Transactions on Robotics}, vol.~40, pp. 3322--3340, 2024.

\bibitem{luo2025bevplace++}
L.~Luo \emph{et~al.}, ``{BEVPlace++}: Fast, robust, and lightweight lidar global localization for autonomous ground vehicles,'' \emph{IEEE Transactions on Robotics}, vol.~41, pp. 4479--4498, 2025.

\bibitem{cui2020acsc}
J.~Cui \emph{et~al.}, ``{ACSC}: Automatic calibration for non-repetitive scanning solid-state {LiDAR} and camera systems,'' \emph{arXiv preprint arXiv:2011.08516}, 2020.

\bibitem{koo2020analytic}
G.~Koo \emph{et~al.}, ``Analytic plane covariances construction for precise planarity-based extrinsic calibration of camera and {LiDAR},'' in \emph{2020 IEEE International Conference on Robotics and Automation (ICRA)}.\hskip 1em plus 0.5em minus 0.4em\relax IEEE, 2020, pp. 6042--6048.

\bibitem{yuan2021pixel}
C.~Yuan \emph{et~al.}, ``Pixel-level extrinsic self calibration of high resolution {LiDAR} and camera in targetless environments,'' \emph{IEEE Robotics and Automation Letters}, vol.~6, no.~4, pp. 7517--7524, 2021.

\bibitem{zhu2020online}
Y.~Zhu \emph{et~al.}, ``Online camera-{LiDAR} calibration with sensor semantic information,'' in \emph{2020 IEEE International Conference on Robotics and Automation (ICRA)}.\hskip 1em plus 0.5em minus 0.4em\relax IEEE, 2020, pp. 4970--4976.

\bibitem{pandey2015automatic}
{G. Pandey} \emph{et~al.}, ``Automatic extrinsic calibration of vision and {LiDAR} by maximizing mutual information,'' \emph{Journal of Field Robotics}, vol.~32, no.~5, pp. 696--722, 2015.

\bibitem{borer2024chaos}
J.~Borer \emph{et~al.}, ``From chaos to calibration: A geometric mutual information approach to target-free camera {LiDAR} extrinsic calibration,'' in \emph{Proceedings of the IEEE/CVF Winter Conference on Applications of Computer Vision (WACV)}, 2024, pp. 8409--8418.

\bibitem{lv2021lccnet}
X.~Lv \emph{et~al.}, ``{LCCNet}: {LiDAR} and camera self-calibration using cost volume network,'' in \emph{Proceedings of the IEEE/CVF Conference on Computer Vision and Pattern Recognition (CVPR)}, 2021, pp. 2894--2901.

\bibitem{shi2020calibrcnn}
J.~Shi \emph{et~al.}, ``{CalibRCNN}: Calibrating camera and {LiDAR} by recurrent convolutional neural network and geometric constraints,'' in \emph{2020 IEEE/RSJ International Conference on Intelligent Robots and Systems (IROS)}.\hskip 1em plus 0.5em minus 0.4em\relax IEEE, 2020, pp. 10\,197--10\,202.

\bibitem{zhao2021calibdnn}
G.~Zhao \emph{et~al.}, ``{CalibDNN}: multimodal sensor calibration for perception using deep neural networks,'' in \emph{Signal Processing, Sensor/Information Fusion, and Target Recognition XXX}, vol. 11756.\hskip 1em plus 0.5em minus 0.4em\relax SPIE, 2021, pp. 324--335.

\bibitem{iyer2018calibnet}
G.~Iyer \emph{et~al.}, ``{CalibNet}: Geometrically supervised extrinsic calibration using {3D} spatial transformer networks,'' in \emph{2018 IEEE/RSJ International Conference on Intelligent Robots and Systems (IROS)}.\hskip 1em plus 0.5em minus 0.4em\relax IEEE, 2018, pp. 1110--1117.

\bibitem{yuan2020rggnet}
K.~Yuan \emph{et~al.}, ``{RGGNet}: Tolerance aware {LiDAR}-camera online calibration with geometric deep learning and generative model,'' \emph{IEEE Robotics and Automation Letters}, vol.~5, no.~4, pp. 6956--6963, 2020.

\bibitem{lv2015automatic}
F.~Lv and K.~Ren, ``Automatic registration of airborne {LiDAR} point cloud data and optical imagery depth map based on line and points features,'' \emph{Infrared Physics \& Technology}, vol.~71, pp. 457--463, 2015.

\bibitem{li2018automatic}
J.~Li \emph{et~al.}, ``Automatic registration of panoramic image sequence and mobile laser scanning data using semantic features,'' \emph{ISPRS Journal of Photogrammetry and Remote Sensing}, vol. 136, pp. 41--57, 2018.

\bibitem{ma2021crlf}
T.~Ma \emph{et~al.}, ``{CRLF}: Automatic calibration and refinement based on line feature for {LiDAR} and camera in road scenes,'' \emph{arXiv preprint arXiv:2103.04558}, 2021.

\bibitem{wang2022automatic}
Y.~Wang \emph{et~al.}, ``Automatic registration of point cloud and panoramic images in urban scenes based on pole matching,'' \emph{International Journal of Applied Earth Observation and Geoinformation}, vol. 115, p. 103083, 2022.

\bibitem{han2021auto}
Y.~Han \emph{et~al.}, ``Auto-calibration method using stop signs for urban autonomous driving applications,'' in \emph{2021 IEEE International Conference on Robotics and Automation (ICRA)}.\hskip 1em plus 0.5em minus 0.4em\relax IEEE, 2021, pp. 13\,179--13\,185.

\bibitem{liao2023se}
Y.~Liao \emph{et~al.}, ``{SE-Calib}: Semantic edges based {LiDAR-camera} boresight online calibration in urban scenes,'' \emph{IEEE Transactions on Geoscience and Remote Sensing}, 2023, {DOI}: 10.1109/TGRS.2023.3278024.

\bibitem{koide2023general}
K.~Koide \emph{et~al.}, ``General, single-shot, target-less, and automatic {LiDAR}-camera extrinsic calibration toolbox,'' in \emph{2023 IEEE International Conference on Robotics and Automation (ICRA)}, 2023, pp. 11\,301--11\,307.

\bibitem{zhiwei2024lcec}
Z.~Huang \emph{et~al.}, ``Online, target-free {LiDAR}-camera extrinsic calibration via cross-modal mask matching,'' \emph{IEEE Transactions on Intelligent Vehicles}, pp. 1--12, 2024.

\bibitem{sarlin2020superglue}
P.-E. Sarlin \emph{et~al.}, ``{SuperGlue}: Learning feature matching with graph neural networks,'' in \emph{Proceedings of the IEEE/CVF Conference on Computer Vision and Pattern Recognition (CVPR)}, 2020, pp. 4938--4947.

\bibitem{schneider2017regnet}
N.~Schneider \emph{et~al.}, ``{RegNet}: Multimodal sensor registration using deep neural networks,'' in \emph{2017 IEEE Intelligent Vehicles Symposium (IV)}.\hskip 1em plus 0.5em minus 0.4em\relax IEEE, 2017, pp. 1803--1810.

\bibitem{zhu2023calibdepth}
J.~Zhu \emph{et~al.}, ``Calibdepth: Unifying depth map representation for iterative {LiDAR}-camera online calibration,'' in \emph{2023 IEEE International Conference on Robotics and Automation (ICRA)}.\hskip 1em plus 0.5em minus 0.4em\relax IEEE, 2023, pp. 726--733.

\bibitem{depth_anything_v2}
L.~Yang \emph{et~al.}, ``{Depth Anything V2},'' \emph{arXiv:2406.09414}, 2024.

\bibitem{wang2025moge}
R.~Wang \emph{et~al.}, ``Moge-2: Accurate monocular geometry with metric scale and sharp details,'' \emph{arXiv preprint arXiv:2507.02546}, 2025.

\bibitem{geiger2012we}
A.~Geiger \emph{et~al.}, ``Are we ready for autonomous driving? the {KITTI} vision benchmark suite,'' in \emph{2012 IEEE conference on Computer Vision and Pattern Recognition (CVPR)}.\hskip 1em plus 0.5em minus 0.4em\relax IEEE, 2012, pp. 3354--3361.

\bibitem{liao2022kitti}
Y.~Liao, J.~Xie, and A.~Geiger, ``Kitti-360: A novel dataset and benchmarks for urban scene understanding in {2D} and {3D},'' \emph{IEEE Transactions on Pattern Analysis and Machine Intelligence}, vol.~45, no.~3, pp. 3292--3310, 2022.

\bibitem{caesar2020nuscenes}
H.~Caesar \emph{et~al.}, ``{nuScenes}: A multimodal dataset for autonomous driving,'' in \emph{Proceedings of the IEEE/CVF Conference on Computer Vision and Pattern Recognition (CVPR)}, 2020, pp. 11\,621--11\,631.

\bibitem{wilson2023argoverse}
B.~Wilson \emph{et~al.}, ``{Argoverse 2}: Next generation datasets for self-driving perception and forecasting,'' \emph{arXiv preprint arXiv:2301.00493}, 2023.

\bibitem{eigen2014depth}
D.~Eigen \emph{et~al.}, ``Depth map prediction from a single image using a multi-scale deep network,'' \emph{Advances in Neural Information Processing Systems (NeurIPS)}, vol.~27, 2014.

\bibitem{godard2019digging}
C.~Godard \emph{et~al.}, ``Digging into self-supervised monocular depth estimation,'' in \emph{Proceedings of the IEEE/CVF International Conference on Computer Vision (ICCV)}, 2019, pp. 3828--3838.

\bibitem{feng2022advancing}
Z.~Feng \emph{et~al.}, ``Advancing self-supervised monocular depth learning with sparse lidar,'' in \emph{Conference on Robot Learning (CoRL)}.\hskip 1em plus 0.5em minus 0.4em\relax PMLR, 2022, pp. 685--694.

\bibitem{huang2025environment}
Z.~Huang \emph{et~al.}, ``Environment-driven online {LiDAR}-camera extrinsic calibration,'' \emph{IEEE Transactions on Automation Science and Engineering}, vol.~22, pp. 24\,164--24\,176, 2025.

\bibitem{yang2018lego}
Z.~Yang \emph{et~al.}, ``{LEGO}: Learning edge with geometry all at once by watching videos,'' in \emph{Proceedings of the IEEE conference on Computer Vision and Pattern Recognition (CVPR)}, 2018, pp. 225--234.

\end{thebibliography}

\clearpage
\clearpage
\setcounter{page}{1}

{
   \newpage
       \twocolumn[
        \centering
        \huge 
        \textbf{
        DST-Calib: A Dual-Path, Self-Supervised, Target-Free LiDAR-Camera Extrinsic Calibration Network
        }\\
        \vspace{0.5em}Supplementary Material \\
        \vspace{1.0em}
       ]
}

\section{Details of DAR}
As introduced in Sect. \ref{sec.method_dar}, in the depth anchors selection of DAR, we employ a two-stage procedure that is robust to outliers and encourages near-linearity while preserving coverage over $d^C$. The pseudo code is presented in Algorithm \ref{alg.two_stage_anchor}\footnote{Since this algorithm is a unified framework that can also be employed for other tasks, we use $x$ and $y$ to replace $d^C$ and $d^L$, respectively.}. Specifically, Stage~I performs $B$ small least-squares fittings and a single pass over the bins, resulting in a computational complexity of $O(N+B)$. Stage~II executes a quadratic dynamic programming (DP) optimization over $M\le B$ candidates, achieving $O(M^2)$ time and $O(M)$ memory complexity. In practice, $M\approx 2T$, thus the overall computational cost remains modest.

\textbf{Stage~I: Coverage-aware candidate thinning.}
Partition the $x$-domain into $B=2\,T$ equal bins, where $T$
is the target number of anchors. From each nonempty bin
$[b_j,b_{j+1}]$ we choose at most one representative
$(\hat x,\hat y)$ that best agrees with the local linear
trend within the bin, e.g., by minimizing a binwise residual
to a least-squares line fit:
\begin{equation}
\label{eq:bin-pick}
(\hat x,\hat y) \in \arg\min_{(x,y)\in \tilde{\mathcal{A}}\cap [b_j,b_{j+1}]}
\bigl|\, y - (\alpha_j x+\beta_j)\,\bigr|,
\end{equation}
where $(\alpha_j,\beta_j)$ is the least-squares solution on the
bin-restricted points.
The resulting candidate set $\mathcal{C}=\{(x_i^{c},y_i^{c})\}_{i=1}^{M}$, $M\le B$,
is $x$-sorted and roughly uniform over depth, which later helps produce
a ``long'' near-linear chain.

\textbf{Stage~II: Dynamic programming for longest monotone convex subsequence.}
Let $\mathcal{C}$ be indexed in increasing $x$.
For each $i$ we store the best valid subsequence $\mathrm{DP}[i]$
ending at $(x_i^{c},y_i^{c})$.
For $i=1,\dots,M$:
\begin{align*}
\mathrm{DP}[i] \leftarrow \{(x_i^{c},y_i^{c})\};\quad
\text{for}\ j<i\ \text{with } y_i^{c}\ge y_j^{c}:
\end{align*}
\[
\text{let } \ s(i,j)=\frac{y_i^{c}-y_j^{c}}{x_i^{c}-x_j^{c}}.
\]
If $|\mathrm{DP}[j]|\le 1$ then we can append $(x_i^{c},y_i^{c})$.
Otherwise, let $(x_{a}^{c},y_{a}^{c})$ be the last point and
$(x_{b}^{c},y_{b}^{c})$ the penultimate point of $\mathrm{DP}[j]$,
and compute the last slope
\[
s_{\text{last}}=\frac{y_{a}^{c}-y_{b}^{c}}{x_{a}^{c}-x_{b}^{c}}.
\]
If $s(i,j)\ge s_{\text{last}}$ (discrete convexity), then update
$$\mathrm{DP}[i]\ \leftarrow\ \arg\max\bigl\{\, \mathrm{DP}[i],\ \mathrm{DP}[j]\cup\{(x_i^{c},y_i^{c})\}\,\bigr\}$$
by cardinality.
Finally, return the longest chain $\mathcal{S}^{\star}=\arg\max_i |\mathrm{DP}[i]|$.
If $|\mathcal{S}^{\star}|>T$, subsample it uniformly in $x$ while retaining
the endpoints to obtain exactly $T$ anchors.

\begin{algorithm}[t!]

\caption{Two-Stage Anchor Selection for Near-Linear Coverage}
\label{alg.two_stage_anchor}
\begin{algorithmic}[1]
\Require Dataset $\tilde{\mathcal{A}} = \{(x_i,y_i)\}_{i=1}^N$, target anchor count $T$
\Ensure Selected anchors $\mathcal{S}$ with $|\mathcal{S}| = T$

\Statex
\textbf{Stage I: Coverage-aware candidate thinning}
\State Partition $x$-domain into $B = 2T$ equal bins
\State Initialize candidate set $\mathcal{C} \leftarrow \emptyset$
\For{each nonempty bin $[b_j, b_{j+1}]$}
    \State Compute least-squares line $(\alpha_j, \beta_j)$ on $\tilde{\mathcal{A}} \cap [b_j,b_{j+1}]$
    \State Select $(\hat x, \hat y) \in \arg\min\limits_{(x,y)\in \tilde{\mathcal{A}}\cap [b_j,b_{j+1}]} | y - (\alpha_j x + \beta_j) |$
    \State $\mathcal{C} \leftarrow \mathcal{C} \cup \{(\hat x, \hat y)\}$
\EndFor
\State Sort $\mathcal{C} = \{(x_i^c,y_i^c)\}_{i=1}^M$ by $x$ (increasing)

\Statex
\textbf{Stage II: Longest monotone convex subsequence}
\State Initialize DP array: $\mathrm{DP}[i] \leftarrow \{(x_i^c,y_i^c)\}$ for $i=1,\dots,M$
\For{$i = 1$ to $M$}
    \For{each $j < i$ with $y_i^c \ge y_j^c$}
        \State Compute slope $s(i,j) \leftarrow \dfrac{y_i^c - y_j^c}{x_i^c - x_j^c}$
        \If{$|\mathrm{DP}[j]| \le 1$}
            \State Candidate $\leftarrow \mathrm{DP}[j] \cup \{(x_i^c,y_i^c)\}$
        \Else
            \State Let $(x_a^c,y_a^c)$ = last point of $\mathrm{DP}[j]$
            \State Let $(x_b^c,y_b^c)$ = penultimate point of $\mathrm{DP}[j]$
            \State $s_{\text{last}} \leftarrow \dfrac{y_a^c - y_b^c}{x_a^c - x_b^c}$
            \If{$s(i,j) \ge s_{\text{last}}$} \Comment{Discrete convexity condition}
                \State Candidate $\leftarrow \mathrm{DP}[j] \cup \{(x_i^c,y_i^c)\}$
            \EndIf
        \EndIf
        \If{Candidate exists and $|\text{Candidate}| > |\mathrm{DP}[i]|$}
            \State $\mathrm{DP}[i] \leftarrow \text{Candidate}$
        \EndIf
    \EndFor
\EndFor
\State $\mathcal{S}^{\star} \leftarrow \arg\max_i |\mathrm{DP}[i]|$

\Statex
\textbf{Post-processing}
\If{$|\mathcal{S}^{\star}| > T$}
    \State Subsample $\mathcal{S}^{\star}$ uniformly in $x$ (retain endpoints) to get exactly $T$ anchors
\EndIf
\State \Return $\mathcal{S}^{\star}$
\end{algorithmic}
\end{algorithm}

\section{Average Calculation of the Relative Pose}
\label{sec.quaternion_rot_avg}
As discussed in Sect \ref{sec.multi_frame_opt}, this article employs a scoring mechanism to 
compute the weighted average of extrinsic parameters in the multi-frame optimization. In this section, we provide the detailed procedure of this weighted averaging process. 

For the translation, the weighted average of the translation vector $\boldsymbol{t}_i$ of $\boldsymbol{T}_i$ is computed as:
\begin{equation}
\boldsymbol{t}^* = \sum_{j=1}^{k} w_j \cdot \boldsymbol{t}_{\pi(j)}.
\end{equation}

Unlike the translation, the weighted average of the rotation matrix cannot be computed by simply taking the mean of the matrices. We use the weighted quaternion averaging via eigenvalue decomposition to calculate the average rotation matrices. First, each rotation component $\boldsymbol{R}_i$ in $\boldsymbol{T}_i$ can be transformed into a quaternion vector $\boldsymbol{q}_i$. The weighted quaternions are constructed by applying the square root of the normalized weights:
\begin{equation}
\tilde{\boldsymbol{q}}_i = \sqrt{\hat{w}_i} \cdot \boldsymbol{q}_i, \quad i = 1, 2, \ldots, n.
\end{equation}
A $4 \times n$ quaternion matrix is formed by concatenating all weighted quaternions $\boldsymbol{Q} = [\tilde{\boldsymbol{q}}_1, \tilde{\boldsymbol{q}}_2, \ldots, \tilde{\boldsymbol{q}}_n]$.
The covariance matrix is computed as:
\begin{equation}
\boldsymbol{C} = \boldsymbol{Q} \boldsymbol{Q}^\top.
\end{equation}
The eigen decomposition of $\mathbf{C}$ yields:
\begin{equation}
\boldsymbol{C} \boldsymbol{e}_j = \lambda_j \boldsymbol{e}_j, \quad j = 1, 2, 3, 4,
\end{equation}
where $\lambda_1 \geq \lambda_2 \geq \lambda_3 \geq \lambda_4$ are the eigenvalues and $\boldsymbol{e}_j$ are the corresponding eigenvectors. Finally, the average quaternion corresponds to the eigenvector associated with the largest eigenvalue $\bar{\boldsymbol{q}} = \boldsymbol{e}_1$. The average quaternion is normalized to ensure unit length:
\begin{equation}
\bar{\boldsymbol{q}} \leftarrow \frac{\bar{\boldsymbol{q}}}{\|\bar{\boldsymbol{q}}\|}.
\end{equation}
Sign consistency is maintained through:
\begin{equation}
\bar{\boldsymbol{q}} \leftarrow 
\begin{cases}
\bar{\boldsymbol{q}}, & \text{if } \bar{\boldsymbol{q}}_0 \geq 0 \\
-\bar{\boldsymbol{q}}, & \text{otherwise}.
\end{cases}
\end{equation}
The corresponding rotation matrix of $\bar{\boldsymbol{q}}$ is the final rotation matrix $\boldsymbol{R}^*$. This multi-frame optimization method can successfully combine calibration results of multiple scenes under the same sensor setup. The weighted average calculation process makes the final estimated extrinsic parameters closer to the real value, thereby improving the overall calibration accuracy.

\section{Details of Datasets}
We have conducted extensive experiments on KITTI odometry, KITTI360, Argoverse2, MIAS-LCEC-TF70, MIAS-LCEC-TF360, nuScenes, and our created LCScenes. Now we provide their details.
\subsubsection{KITTI Odometry} 
KITTI Odometry is a large-scale public dataset recorded using a vehicle equipped with two RGB cameras Point Grey Flea 2, and one LiDAR of type Velodyne HDL-64E. Sensor data is captured at 10 Hz. The first 10 sequences (00-09) are utilized for evaluation. Specifically, the data from the 08 sequence are used for training, and the others are used for testing.
\subsubsection{KITTI-360} 
KITTI-360 is a large-scale autonomous driving dataset designed to advance research across semantic scene understanding, 3D object detection, and SLAM. We use images captured by the front stereo cameras and point clouds from the Velodyne HDL-64E LiDAR. The first 3000 pairs of LiDAR point clouds and camera images in the 00 sequence are used to evaluate DST-Calib's zero-shot generalization performance.
\subsubsection{Argoverse2} 
Argoverse2 is a large-scale, multimodal dataset designed for advancing perception and forecasting research in autonomous driving. It provides high-quality sensor data collected across diverse urban environments in the United States. The recording vehicle is equipped with two Velodyne VLP-32 LiDARs, seven ring RGB cameras (1920×1200 at 30 FPS), and two forward-facing stereo cameras (2056×2464 at 5 FPS). Since it includes many different LiDAR-camera sensor configurations, it is well-suited for validating the generalization ability of LCEC methods. The ring camera in the front center view and the ring camera in the front left view are used for the training of DST-Calib in the cross-domain generalization experiment.
\subsubsection{MIAS-LCEC-TF70}
MIAS-LCEC-TF70 is constructed as a challenging multi-modal dataset and includes 60 paired samples of 4D point clouds and RGB images. The point clouds record both 3D spatial coordinates and intensity values. Data collection was carried out using a Livox Mid-70 LiDAR together with a MindVision SUA202GC camera. The dataset spans a broad range of indoor and outdoor scenes and was acquired under diverse operating conditions, including different environments, weather patterns, and lighting variations, thereby increasing its overall complexity and diversity.
\subsubsection{MIAS-LCEC-TF360}
The MIAS-LCEC-TF360 dataset consists of 12 pairs of 4D point clouds and RGB images captured in both indoor and outdoor scenarios using a Livox Mid-360 LiDAR and a MindVision SUA202GC camera. Due to the full 360$^\circ$ scanning capability of the Mid-360 sensor, the generated point clouds are considerably sparser than those obtained with the Mid-70 LiDAR. In addition, the pronounced mismatch in field of view between the LiDAR and the camera results in a relatively limited overlap between the two modalities. These characteristics make MIAS-LCEC-TF360 particularly suitable for evaluating algorithm performance in scenarios with sparse data and minimal cross-modal overlap.
\subsubsection{nuScenes}
The LiDAR used in nuScenes is a 32-line sensor HDL-32E, which is significantly sparser than the 64-line LiDAR employed in KITTI and KITTI-360, making the calibration task more challenging. We use the official nuScenes SDK to generate image-point cloud pairs from the 150 testing scenes to construct our evaluation sets. 
\subsubsection{LCScenes}
LCScenes is a dataset recorded using our customized device. The LiDAR and camera are installed on an adjustable platform placed on a tripod. It contains extensive pairs of 4D point clouds and 2D camera images with different extrinsic parameters, captured in different indoor and outdoor scenarios. The LiDAR sensor is Livox-Mid70.  The resolution of the camera sensors is 1200 × 800 and 2400 × 1200. This dataset is divided into five sequences, each recorded along a specific route.

\end{document}